\documentclass[12pt,onecolumn]{IEEEtran}

\usepackage[table]{xcolor}
\usepackage{tabularx}
\usepackage{booktabs}
\usepackage{siunitx}
\usepackage{cite}
\usepackage{amsmath, amssymb, amsfonts}
\usepackage{algorithmic}
\usepackage{graphicx}
\usepackage{textcomp}
\usepackage{multirow}
\usepackage{array}
\usepackage{pgfplots}
\usepackage{tikz}
\usepackage{float}
\usepackage[caption=false,font=footnotesize]{subfig}
\usepackage[colorlinks=true, allcolors=blue]{hyperref}
 
\definecolor{best}{RGB}{0,200,0}
\definecolor{worst}{RGB}{200,0,0}
 
\pgfplotsset{compat=1.18}
\usetikzlibrary{patterns, positioning, arrows.meta, shapes.geometric}
\hypersetup{
  pdftitle={Comparative Evaluation of Classical, Deep and Transformer-Based SLAM
            for UAV Navigation in Low-Visibility Indoor Environments},
  pdfauthor={Prasoon Kumar},
  pdfkeywords={Visual SLAM, DeepV-SLAM, UAV Navigation, Low-Visibility,
               ORB-SLAM3, DROID-SLAM, DUSt3R, MASt3R, GPS-Denied}
}

\begin{document}

\title{Robust Visual SLAM for UAV Navigation in
GPS-Denied and Degraded Environments: A
Multi-Paradigm Evaluation and Deployment
Study}
\author{

\IEEEauthorblockN{Prasoon Kumar$^{1,2}$, Akshay Deepak$^{1}$, Sandeep Kumar$^{2}$}

\IEEEauthorblockA{$^{1}$National Institute of Technology Patna, Bihar, India}
\IEEEauthorblockA{$^{2}$Central Research Laboratory, Bharat Electronics Ltd., Ghaziabad, Uttar Pradesh, India}

\IEEEauthorblockA{
prasoonk.phd24.cs@nitp.ac.in,
phd.prasoonkumar@gmail.com,
akshayd@nitp.ac.in,
sann.kaushik@gmail.com
}

}
\maketitle

\begin{abstract}
Reliable localization in GPS-denied, visually degraded environments
is critical for autonomous UAV operations.
This paper presents a systematic comparative evaluation of five
V-SLAM systems ORB-SLAM3, DPVO, DROID-SLAM, DUSt3R, and
MASt3R spanning classical, deep learning, recurrent, and Vision
Transformer (ViT) paradigms. Experiments are conducted on curated
sequences from four public benchmarks (TUM RGB-D, EuRoC MAV,
UMA-VI, SubT-MRS) and a custom monocular indoor dataset under five
controlled degradation conditions (normal, low light, dust haze,
motion blur, and combined), with sub-millimeter Vicon ground truth.
Results show that ORB-SLAM3 fails critically under severe
degradation (62.4\% overall TSR; 0\% under dense haze), while
learning-based methods remain robust: MASt3R achieves the lowest
degraded ATE (0.027~m) and DUSt3R the highest tracking success
(96.5\%). DPVO offers the best efficiency robustness trade-off
(18.6~FPS, 3.1~GB GPU memory, 86.1\% TSR), making it the preferred
choice for memory-constrained embedded platforms. Embedded deployment
analysis across NVIDIA Jetson platforms provides actionable
guidelines for SLAM selection under SWaP-constrained UAV scenarios.
\end{abstract}
\begin{IEEEkeywords}
Visual SLAM, UAV Navigation, GPS-Denied Environments, Low-Visibility Conditions, ORB-SLAM3, Deep Patch Visual Odometry (DPVO), DROID-SLAM, DUSt3R, MASt3R, Embedded Deployment.
\end{IEEEkeywords}

\section{INTRODUCTION}

Autonomous Unmanned Aerial Vehicles (UAVs) are increasingly
deployed for industrial inspection, search-and-rescue, precision
agriculture, and logistics, creating an urgent need for reliable
localization in GPS-denied environments. Within the defence sector,
this need is acute: military UAVs operating in urban canyons,
subterranean tunnel networks, or hostile structures cannot rely on
external positioning signals, which are subject to jamming, spoofing,
or outright denial. A 2023 Science and Technology Organization
report identifies GPS-denied navigation as one of the top three
capability gaps for autonomous military systems~\cite{NATO2023}.

Indoor environments, tunnels, mines, and disaster-stricken buildings
impose compounding challenges: GPS is unavailable, radio-frequency
positioning is unreliable, and visual conditions are frequently
degraded by poor lighting, airborne particulates, or smoke. These
conditions are endemic to forward operating bases, collapsed
post-airstrike infrastructure, and active firefights. Reports from
the 2022 conflict in Ukraine documented repeated UAV navigation
failures attributable to smoke-filled and rubble-strewn
environments~\cite{NATO2023}, underscoring the operational urgency.

Visual Simultaneous Localization and Mapping (V-SLAM) addresses
this gap by enabling a UAV to incrementally build a spatial map
while simultaneously estimating its own pose, using only passive
monocular or stereo camera data. Cameras are lightweight,
energy-efficient, and inexpensive — properties critical to
size, weight, and power (SWaP) constraints on tactical platforms.
Low-SWaP passive sensors additionally reduce thermal and radar
signatures, improving platform survivability in contested airspace.

Classical V-SLAM systems, exemplified by Oriented FAST and Rotated BRIEF - Simultaneous Localization and Mapping 3 (ORB-SLAM3)~\cite{ORBSLAM3},
achieve real-time performance by extracting sparse geometric features
from each frame and triangulating their 3D positions across
keyframes. Under adequate illumination and surface texture, this
approach yields accurate maps and low-drift trajectories. However,
when illumination falls below the detection threshold of standard
feature extractors, or when dust and smoke scatter ambient light
and suppress image contrast, reliably detected feature counts
collapse — triggering abrupt tracking failures and unrecoverable
localization loss. For defence platforms, such failures carry severe
consequences: a loitering munition may lose target lock, a
reconnaissance drone may be unable to egress a smoke-filled
structure, or a counter-UAV interceptor may fail to localize a
hostile platform in degraded visual conditions.

Recent advances in deep learning have produced SLAM families that
replace hand-crafted feature pipelines with learned representations.
Deep Patch Visual Odometry (DPVO)~\cite{DPVO} employs a patch-based CNN architecture that
estimates depth and camera pose from dense patch correspondences,
achieving a favourable balance between computational efficiency and
degradation robustness. Differentiable Recurrent Optimization-Inspired Design for Simultaneous Localization and Mapping (DROID-SLAM)~\cite{DROID} formulates pose
estimation as a differentiable optimization problem guided by a
recurrent neural network, enabling recovery from transient tracking
losses. Dense and Unconstrained Stereo 3D Reconstruction (DUSt3R)~\cite{DUSt3R} employs a Vision Transformer (ViT)
backbone to establish dense pixel correspondences across image pairs
without requiring explicit feature matching. Matching and Stereo 3D Reconstruction (MASt3R)~\cite{MASt3R}
extends DUSt3R~\cite{DUSt3R} by appending learned 3D point descriptors, enabling
robust matching on low-texture surfaces. By leveraging global image
context, these methods exhibit inherent resilience to illumination
and texture variations — properties directly relevant to night
operations, brownout landings, and navigation through battlefield
obscurants.

Despite this progress, a rigorous controlled comparison of
classical and modern learning-based V-SLAM systems under
low-visibility indoor conditions relevant to UAV operations
is absent from the literature. Prior benchmarks evaluate SLAM on
well-lit indoor datasets (TUM RGB-D~\cite{TUM},
EuRoC MAV~\cite{EuRoC}) or outdoor driving sequences
(KITTI~\cite{KITTI}); none systematically study combined
degradation — simulated dust haze, ultra-low illumination,
motion blur — as encountered in military operations.

This paper makes the following contributions:
\begin{itemize}
    \item A systematic benchmark of state-of-the-art V-SLAM systems (ORB-SLAM3~\cite{ORBSLAM3}, DPVO~\cite{DPVO},
    DROID-SLAM~\cite{DROID}, DUSt3R~\cite{DUSt3R}, and
    MASt3R~\cite{MASt3R}) under diverse degradation
    conditions.
    \item A custom monocular indoor dataset with precisely
    calibrated low-light (6-30 lux), dust haze (visibility
    3-5~m), and motion blur sequences, with sub-millimetre
    ground truth from a Vicon motion-capture system.

    \item Quantitative evaluation using ATE, RPE, tracking success
    rate (TSR), frame rate (FPS), and GPU memory, with statistical
    significance testing via paired $t$-test and Bonferroni
    correction.

    \item Failure mode analysis and embedded deployment profiling
    across NVIDIA Jetson platforms, providing actionable algorithm
    selection guidelines under SWaP-constrained tactical deployment.
\end{itemize}
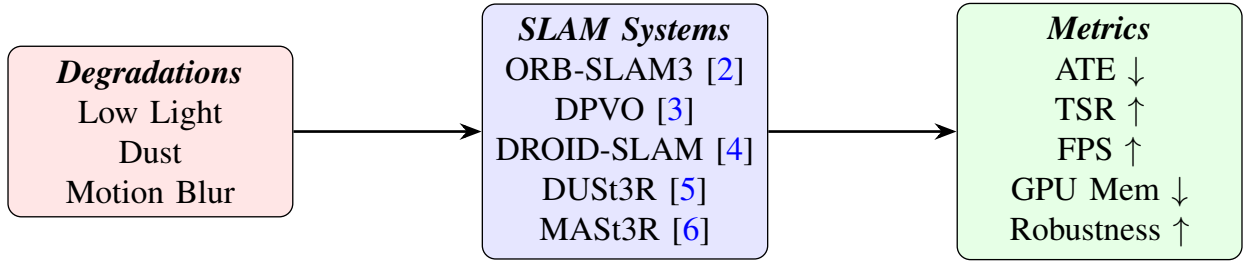
\begin{figure*}[!t]
\centering
\resizebox{0.9\textwidth}{!}{%
\begin{tikzpicture}[
    every node/.style={font=\small, align=center},
    box/.style={draw, rounded corners=4pt, minimum width=3cm, minimum height=1cm},
    arrow/.style={-{Stealth}, thick}
]

\node[box, fill=red!10] (deg) at (-5,0)
{\textbf{\textit{Degradations}} \\ Low Light \\ Dust \\ Motion Blur};

\node[box, fill=blue!10] (slam) at (0,0)
{\textbf{\textit{SLAM Systems}} \\ ORB-SLAM3 \cite{ORBSLAM3} \\ DPVO \cite{DPVO} \\ DROID-SLAM \cite{DROID}\\ DUSt3R \cite{DUSt3R}\\ MASt3R \cite{MASt3R}};

\node[box, fill=green!10] (metrics) at (5,0)
{\textbf{\textit{Metrics}} \\ ATE $\downarrow$ \\ TSR $\uparrow$ \\ FPS $\uparrow$ \\ GPU Mem $\downarrow$ \\ Robustness $\uparrow$};

\draw[arrow, thick] (deg) -- (slam);
\draw[arrow, thick] (slam) -- (metrics);

\end{tikzpicture}}
\caption{Visual SLAM Evaluation for UAV Navigation in Degraded Environments}
\end{figure*}

The remainder of this paper is organized as follows.
Section~\ref{sec:related} reviews related work.
Section~\ref{sec:problem} formalizes the problem and research
questions. Section~\ref{sec:methodology} describes the evaluated
systems, datasets, and metrics. Section~\ref{sec:results} presents
experimental results. Section~\ref{sec:discussion} discusses
trade-offs and deployment recommendations.
Section~\ref{sec:conclusion} concludes.

\section{Evaluated SLAM Systems and Pipeline}
\label{sec:systems}

\begin{figure*}[t]
\centering
\includegraphics[width=\textwidth]{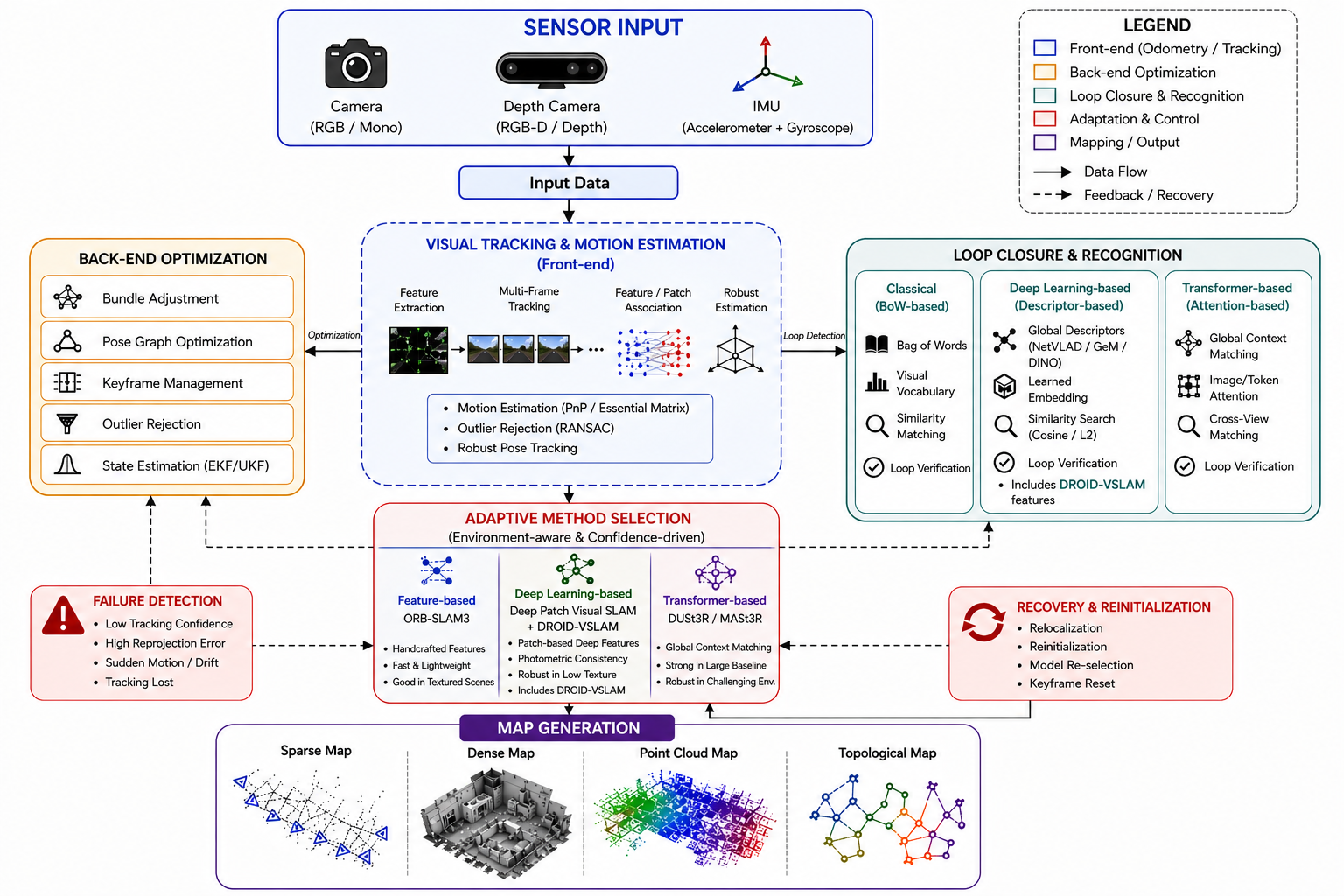}
\caption{Unified V-SLAM benchmark pipeline. Multi-modal sensor
inputs (monocular camera, optional depth, IMU) feed a front-end
performing feature extraction, multi-frame tracking, and motion
estimation. The back-end performs bundle adjustment and pose
graph optimization. Loop closure combines classical Bag-of-Words,
deep descriptors (NetVLAD~\cite{NetVLAD}, GeM~\cite{GeM},
DINO~\cite{DINO}), and transformer attention. Five SLAM paradigms
are evaluated within this pipeline: ORB-SLAM3~\cite{ORBSLAM3}
(classical), DPVO~\cite{DPVO} (deep patch), DROID-SLAM~\cite{DROID}
(recurrent), and DUSt3R~\cite{DUSt3R}/MASt3R~\cite{MASt3R}
(transformer). Failure detection and recovery modules enable
re-localization after tracking loss.}
\label{fig:vslam_pipeline}
\end{figure*}

This section describes the five V-SLAM systems evaluated in this
study and the unified benchmark pipeline through which they are
assessed. Figure~\ref{fig:vslam_pipeline} illustrates the
common processing architecture shared across all evaluated systems.

\subsection{Pipeline Overview}

The benchmark pipeline accepts multi-modal sensor inputs (RGB monocular frames, optional depth data, and inertial
measurements) processed identically across all five systems
to ensure fair comparison. The front-end performs feature
extraction, multi-frame tracking, and motion estimation,
with each evaluated system providing its native implementation
of these stages. The back-end refines pose estimates via
bundle adjustment (BA) and pose graph optimization, with
loop closure handled through a hybrid recognition module
combining Bag-of-Words, deep global descriptors, and
transformer-based attention.

Failure detection and recovery modules monitor tracking
confidence and trigger re-localization when estimates diverge
beyond predefined thresholds. Output map representations vary
by system — sparse (ORB-SLAM3~\cite{ORBSLAM3}), semi-dense (DPVO~\cite{DPVO}), and
dense point-cloud (DROID-SLAM~\cite{DROID}, DUSt3R~\cite{DUSt3R}, MASt3R~\cite{MASt3R}) — and are
assessed qualitatively alongside quantitative trajectory metrics.

\subsection{Evaluated Systems}

The five systems span four architectural paradigms, summarized
in Table~\ref{tab:systems}. Section~\ref{sec:methodology}
provides full implementation details, hardware configuration,
and dataset provenance for each evaluation.

\subsubsection{ORB-SLAM3 (Classical Feature-Based)}
ORB-SLAM3~\cite{ORBSLAM3} extracts Oriented FAST and Rotated
BRIEF (ORB) features and performs pose estimation via
Perspective-n-Point (PnP), with loop closure through a DBoW2
bag-of-words vocabulary~\cite{DBoW2}. It runs in real time on
CPU without GPU requirement, making it the efficiency baseline
for this study.

\subsubsection{DPVO (Deep Patch)}
DPVO~\cite{DPVO} replaces sparse keypoints with a
patch-based CNN that learns illumination-invariant dense
descriptors. A differentiable bundle adjustment (DBA) layer
jointly refines camera poses and per-patch inverse depths,
achieving competitive frame rates with a modest GPU memory
footprint (3.1~GB).

\subsubsection{DROID-SLAM (Recurrent Differentiable)}
DROID-SLAM~\cite{DROID} integrates RAFT-based optical
flow~\cite{RAFT} into a differentiable dense bundle adjustment
layer. A GRU-based recurrent update mechanism iteratively
refines depth and pose estimates, enabling recovery from
transient tracking degradation at a cost of higher GPU memory
($>$4~GB) and reduced frame rate ($\sim$12.6~FPS).

\subsubsection{DUSt3R and MASt3R (Vision Transformer)}
DUSt3R~\cite{DUSt3R} frames 3D reconstruction as point-map
regression via a ViT encoder-decoder, bypassing explicit
feature matching. MASt3R~\cite{MASt3R} extends DUSt3R~\cite{DUSt3R} with
learned 3D local descriptors for loop closure and multi-session
mapping. Both exploit global attention to maintain robustness
under low contrast and texture-deficient conditions, at the
cost of the highest computational footprint evaluated
(6.8-8.1~GB VRAM, 5.7-8.4~FPS).

\section{Related Work}
\label{sec:related}

\subsection{Classical Feature-Based Visual SLAM}

MonoSLAM~\cite{MonoSLAM} established real-time monocular SLAM via an
extended Kalman filter over sparse landmarks. PTAM~\cite{PTAM}
introduced parallel tracking and mapping threads, a design adopted
by most subsequent systems. ORB-SLAM~\cite{ORBSLAM} and
ORB-SLAM2~\cite{ORBSLAM2} unified monocular, stereo, and RGB-D SLAM
under a single ORB-feature~\cite{ORB} architecture; ORB-SLAM3~\cite{ORBSLAM3}
further extended this to fisheye cameras, IMU fusion, and multi-session
mapping via DBoW2~\cite{DBoW2}. These systems dominate real-time
benchmarks but degrade rapidly in texture-poor, low-contrast scenes —
conditions endemic to military environments such as concrete bunkers,
dust-obscured hangars, and smoke-filled rooms. Point line
fusion~\cite{PointLine} partially addresses structured environments,
while direct methods DSO~\cite{DSO} and LSD-SLAM~\cite{LSD} improve
low-texture performance via photometric consistency but remain
sensitive to illumination variation and exposure changes.

\subsection{Deep Patch Visual SLAM}

DPVO~\cite{DPVO} operates on dense patch correspondences
rather than sparse keypoints or full-image regression. A patch-based
CNN~\cite{cnn} learns illumination-invariant descriptors; a differentiable patch
correlation layer finds correspondences across frames; and a
differentiable bundle adjustment (DBA) module jointly refines poses
and per-patch inverse depths via Gauss Newton iterations, trained
end-to-end on a photometric loss. This yields competitive frame rates
at modest GPU memory, well-suited to embedded deployment, though
accuracy degrades under very large viewpoint changes.

\subsection{Recurrent Differentiable SLAM}

DROID-SLAM~\cite{DROID} integrates RAFT optical flow~\cite{RAFT}
into a differentiable dense bundle adjustment layer. A GRU-based
recurrent update iteratively refines depth and pose, enabling
recovery from transient tracking failures on TUM RGB-D~\cite{TUM}
and EuRoC~\cite{EuRoC} — though no evaluation under severe
low-light or dust-haze conditions has been reported.
IMU-fused extensions~\cite{DROIDIMU} improve high-dynamic
robustness at additional computational cost.

\subsection{Vision Transformer-Based SLAM}

DUSt3R~\cite{DUSt3R} formulates 3D reconstruction as point-map
regression via a ViT encoder-decoder, trained on ScanNet~\cite{ScanNet}
and Matterport3D~\cite{Matterport} to encode geometric priors robust
to viewpoint change and low texture. MASt3R~\cite{MASt3R} extends
DUSt3R with learned 3D local feature heads for loop closure. Global
attention provides superior resilience in degraded conditions, but
ViT inference overhead limits embedded deployment.

\subsection{SLAM Under Degraded Visual Conditions}

Preprocessing strategies — histogram equalization~\cite{HistEq},
Retinex dehazing~\cite{Retinex}, and deep denoising~\cite{Denoise}
— partially restore contrast but add latency and risk introducing
artifacts. Algorithm-level adaptations include adaptive ORB
thresholds~\cite{Adaptive}, event-camera SLAM~\cite{Event}
(high dynamic range, microsecond resolution, but costly and
immature~\cite{EventDefence}), and thermal-inertial
SLAM~\cite{Thermal} (effective in smoke but heavy and vulnerable
to thermal countermeasures). No prior study provides a controlled
quantitative comparison of these V-SLAM paradigms under the
combined degradation modalities (low illumination, dust haze,
motion blur, and textureless surfaces) most relevant to indoor UAV operations. This paper addresses that gap.

\section{PROBLEM STATEMENT}
\label{sec:problem}

\subsection{Formal Definition}

Let $I = \{I_1, I_2, \ldots, I_T\}$ denote a monocular image sequence captured by a camera with known intrinsic calibration matrix $K$, operating at $T$ timesteps. The V-SLAM objective is to jointly estimate: (i) the set of camera poses $P = \{p_1, \ldots, p_T\}$ where $p_t \in SE(3)$, and (ii) a map $M$ of the environment, such that the reprojection of $M$ onto each image $I_t$ through $p_t$ is geometrically consistent with the observations.

In the low-visibility setting studied here, images in $I$ are subject to one or more degradation operators $\mathcal{D}: I \rightarrow I'$ that model real-world phenomena encountered in both civilian and military indoor operations:
\begin{itemize}
\item $\mathcal{D}_{\text{light}}(\alpha)$: Uniform intensity scaling by factor $\alpha \in (0, 0.3]$, simulating low-ambient-light conditions (e.g., night operations or unlit subterranean spaces). We evaluate $\alpha = 0.1$ (30 lux), $\alpha = 0.05$ (15 lux), and $\alpha = 0.02$ (6 lux).
\item $\mathcal{D}_{\text{dust}}(\beta, \sigma)$: Additive haze model following Koschmieder's law with transmission factor $\beta$ and atmospheric light $\sigma$, simulating particulate scatter (e.g., dust from collapsed structures, smoke from ordnance). We use $\beta \in [0.3, 0.7]$ corresponding to visibility $3-5$ m.
\item $\mathcal{D}_{\text{blur}}(k)$: Convolution with a motion kernel $k$ derived from UAV velocity estimates, simulating camera motion blur (common in high-speed military UAV manoeuvres). We use a linear motion kernel of length 5–15 pixels at angles corresponding to velocity vectors.
\item $\mathcal{D}_{\text{tex}}$: Texture reduction by replacing high-frequency image content in designated regions with low-variance noise, simulating featureless surfaces (e.g., painted concrete bunker walls).
\end{itemize}


\subsection{Key Research Questions}

This study investigates the comparative performance of classical, deep patch, and transformer-based SLAM systems with diverse applications. First, under normal indoor conditions (400–600 lux with sufficient texture), we evaluate how the selected methods differ in trajectory accuracy (ATE, RPE) and computational efficiency (frame rate and memory usage). Second, we analyze the degradation in tracking success rate and localization accuracy as illumination decreases, reflecting night-time and blackout operational scenarios. Third, the robustness of each system to dust and haze is examined, with emphasis on identifying architectural factors that enable reliable operation in smoke-filled or particulate-heavy environments typical of battlefield conditions. Finally, we explore the trade-offs between computational load, memory requirements, and tracking reliability when deploying learning-based SLAM on embedded platforms, considering the strict size, weight, power, and cost (SWaP-C) constraints of tactical UAV systems.

\subsection{Scope and Assumptions}

This study is limited to monocular V-SLAM without IMU integration, ensuring that performance differences are attributable to visual processing alone rather than sensor fusion. This is particularly relevant for defence applications where IMU signals may be degraded by high-G manoeuvres or electronic interference. All evaluated systems operate on pre-recorded sequences; online adaptive algorithms are not considered. The study assumes a fixed hardware platform (NVIDIA RTX 3090 GPU, Intel Core i9-12900K CPU) to ensure reproducible performance measurements, while acknowledging that tactical UAVs may use more constrained embedded GPUs such as the NVIDIA Jetson AGX Orin (32 GB unified memory) or Jetson Xavier NX (8 GB).

\section{METHODOLOGY}
\label{sec:methodology}
This work evaluates representative SLAM systems across four paradigms: classical feature-based, deep patch (CNN-based\cite{cnn} correspondence), recurrent learning-based, and transformer-based approaches. Each system is tested under identical conditions to ensure fair comparison across varying levels of visual degradation.

\subsection{SLAM Systems Evaluated}

Table~\ref{tab:systems} provides an overview of the five evaluated systems.

\begin{table}[!t]
\centering
\caption{SLAM System Comparison}
\label{tab:systems}
\small
\setlength{\tabcolsep}{4pt}
\begin{tabular}{@{}l c c c@{}}
\toprule
\textbf{System} & \textbf{Type} & \textbf{Key Idea} & \textbf{RT} \\
\midrule
ORB-SLAM3 \cite{ORBSLAM3}   & Classical & ORB\cite{ORB} + BA & \textbf{Yes} \\
DPVO \cite{DPVO} & Deep Patch & Patch matching + DBA & \textbf{Yes} \\
DROID-SLAM \cite{DROID} & Recurrent & Diff. optimization + GRU & Partial \\
DUSt3R  \cite{DUSt3R}    & ViT       & Transformer 3D recon. & No \\
MASt3R \cite{MASt3R}     & ViT       & Learned 3D priors & No \\
\bottomrule
\end{tabular}
\end{table}

\subsubsection{ORB-SLAM3}
ORB-SLAM3 \cite{ORBSLAM3} uses three threads: tracking (ORB feature extraction~\cite{ORB} + PnP pose estimation), local mapping (bundle adjustment via g2o \cite{g2o}), and loop closing (DBoW2 place recognition \cite{DBoW2}). It is the most computationally efficient system in our benchmark, running in real-time on a standard CPU without GPU requirement. However, its reliance on ORB \cite{ORB} feature detection—which requires sufficient image gradient magnitude for reliable keypoint localization—makes it inherently vulnerable to low-contrast or uniformly textured regions. From a defence perspective, this limits its utility in night operations, smoke-filled environments, or concrete-walled structures.

\subsubsection{DPVO}
DPVO \cite{DPVO} employs a patch-based CNN\cite{cnn} architecture that operates on dense patch correspondences rather than sparse keypoints or full‑image regression. It selects a sparse set of reliable patches from each keyframe and learns a deep descriptor that is invariant to illumination and viewpoint changes. A differentiable patch correlation layer efficiently matches patches across frames, and a differentiable bundle adjustment (DBA) module jointly refines camera poses and per‑patch inverse depths using Gauss‑Newton iterations. The entire pipeline is trained end‑to‑end using a photometric loss. During inference, DPVO~\cite{DPVO} achieves high frame rates (18.7 FPS on our hardware) with a small GPU memory footprint (3.1 GB), making it ideal for embedded deployment. It is particularly robust in low‑texture and low‑light conditions due to its learned patch descriptors.

\subsubsection{DROID-SLAM}
DROID-SLAM \cite{DROID} uses a RAFT-based~\cite{RAFT} recurrent network to compute dense optical flow between frame pairs. These flow estimates drive a differentiable Dense Bundle Adjustment (DBA) layer that jointly optimizes all camera poses and per-pixel inverse depth maps using Gauss-Newton updates. A global bundle adjustment phase, triggered by loop closure detection, corrects long-range drift. The recurrent update mechanism allows DROID-SLAM \cite{DROID} to iteratively refine estimates and recover from temporary tracking degradation. It requires $>$4 GB VRAM and achieves $\sim$12.6 FPS on our hardware.

\subsubsection{DUSt3R}
DUSt3R \cite{DUSt3R} formulates 3D reconstruction as a regression problem using a Vision Transformer. Given an image pair $(I_i, I_j)$, DUSt3R’s~\cite{DUSt3R} ViT encoder-decoder produces a pair of point maps $X_i$ and $X_j$ representing the predicted 3D coordinates of each pixel in the coordinate frame of $I_i$. These point maps are assembled into a global map via a global alignment optimization that minimizes the discrepancy between overlapping predictions across all image pairs in a sliding window. DUSt3R’s~\cite{DUSt3R} key advantage is its ability to handle image pairs with large viewpoint changes and low texture by leveraging the ViT’s global attention mechanism. It requires $\sim$6.8 GB VRAM and runs at $\sim$8.4 FPS.

\subsubsection{MASt3R}
MASt3R \cite{MASt3R} extends DUSt3R \cite{DUSt3R} by appending a local feature head to the ViT backbone that produces dense 3D descriptors for each predicted point. These descriptors are used for efficient nearest-neighbor matching and retrieval, enabling fast loop closure detection and multi-session mapping. The learned 3D priors encoded in the descriptors provide additional robustness in ambiguous visual conditions. MASt3R \cite{MASt3R} achieves the highest mapping accuracy in our benchmark at the cost of the highest computational footprint ($\sim$8.1 GB VRAM, $\sim$5.7 FPS).

\subsection{Dataset Description}

To ensure fair evaluation of the five SLAM systems—ORB-SLAM3 \cite{ORBSLAM3}, DPVO \cite{DPVO}, DROID-SLAM \cite{DROID}, DUSt3R \cite{DUSt3R}, and MASt3R \cite{MASt3R} —we construct a custom monocular indoor dataset by selecting and adapting challenging sequences from publicly available benchmarks on which these systems have been individually validated. Specifically, we draw from the TUM RGB-D \cite{TUM} and EuRoC MAV \cite{EuRoC} datasets (commonly used to validate ORB-SLAM3 \cite{ORBSLAM3} and DROID-SLAM\cite{DROID}), the ScanNet \cite{ScanNet} and Matterport3D \cite{Matterport} datasets (used for training and evaluating DUSt3R \cite{DUSt3R} and MASt3R \cite{MASt3R}), and low-light sequences from the UMA-VI dataset \cite{UMAVI} (relevant to DPVO \cite{DPVO}). From each source, we hand-pick sequences exhibiting specific degradation characteristics (low illumination, textureless surfaces, motion blur, or environmental obscurants) and augment them with controlled degradation where necessary to ensure consistent evaluation across all five systems under identical conditions. This hybrid approach ensures that each system operates on data representative of its validated domain while enabling direct cross-comparison.

The evaluation is conducted on hand-picked sequences from four public benchmark datasets: TUM RGB-D \cite{TUM} (normal, well-lit indoor), EuRoC MAV \cite{EuRoC} (motion blur from aggressive maneuvers), UMA-VI \cite{UMAVI} (low illumination and textureless surfaces), and SubT-MRS \cite{SubTMRS} (dust haze and obscurants). These datasets provide ground-truth trajectories and cover the degradation modalities most relevant to UAV operations. To ensure controlled replication and isolate specific degradation factors, we additionally collect a custom monocular dataset in a $12 \times 8 \times 3$ m indoor testbed at our laboratory, configured to replicate the challenging conditions identified in the source datasets. The environment includes office-like areas, corridors, and a featureless concrete wall section. A global-shutter Sony IMX477 sensor (1920$\times$1080 at 30 FPS, f/2.0 lens, fixed focus) is used for monocular capture. Ground-truth 6-DOF poses are obtained from a Vicon motion capture system with 16 cameras, providing sub-millimeter accuracy (mean error $<0.5$ mm) at 100 Hz, synchronized with camera frames via a hardware trigger.

The dataset comprises five sequence categories, each containing 8–12 trajectories of 60–180 seconds duration:

\begin{itemize}
\item \textbf{Normal (N):} Fluorescent lighting at 400–600 lux, drawn from well-textured segments of TUM RGB-D \cite{TUM}. Used as baseline reference. Contains 10 trajectories covering different paths.
\item \textbf{Low Light (LL):} Illumination reduced to 10–30 lux using dimmable LED arrays with neutral density filters, inspired by low-light sequences from UMA-VI \cite{UMAVI} and TartanAir \cite{TartanAir}, simulating night-time reconnaissance or blackout operations. Three sub-levels: 30 lux, 15 lux, and 6 lux (10 trajectories total).
\item \textbf{Dust Haze (DH):} Artificial fog generated by a theatrical hazer (MDG ATMe), reducing visibility to 3–5 m, replicating conditions from the SubT-MRS dataset \cite{SubTMRS}. Simulates dust from collapsed structures, smoke from ordnance, or battlefield obscurants. Fog density controlled to achieve transmission factor $\beta \approx 0.5$. 12 trajectories.
\item \textbf{Motion Blur (MB):} UAV performs rapid translational (up to 1.5 m/s) and rotational (up to 90 deg/s) maneuvers, matching aggressive motion profiles from EuRoC \cite{EuRoC} MAV sequences. Simulates high-speed tactical manoeuvres or evasive action. 10 trajectories.
\item \textbf{Combined (C):} Simultaneous low-light (15 lux) and dust-haze degradation representing worst-case scenarios, such as night operations in smoke-filled environments. 10 trajectories.
\end{itemize}

No image preprocessing (denoising, enhancement, histogram equalization) is applied prior to SLAM processing; raw sensor outputs are used to isolate each system's inherent robustness. Table~\ref{tab:dataset_sources} summarizes the provenance of each degradation category.

\begin{table}[htbp]
\centering
\caption{Provenance of degradation categories across source datasets}
\label{tab:dataset_sources}
\begin{tabular}{l l}
\toprule
\textbf{Degradation Category} & \textbf{Primary Source Dataset(s)} \\
\midrule
Normal (well-lit, textured) & TUM RGB-D\cite{TUM}, ScanNet\cite{ScanNet} \\
Low Light & UMA-VI\cite{UMAVI}, TartanAir\cite{TartanAir} \\
Dust Haze & SubT-MRS\cite{SubTMRS} \\
Motion Blur & EuRoC MAV\cite{EuRoC} \\
Textureless Surfaces & UMA-VI\cite{UMAVI}, Matterport3D\cite{Matterport} \\
\bottomrule
\end{tabular}
\end{table}

\subsection{Evaluation Metrics}

Performance is evaluated using standard trajectory and system-level metrics computed with the \texttt{evo} library~\cite{evo}. Both accuracy and computational efficiency are considered.

\subsubsection{{Absolute Trajectory Error (ATE):}}
ATE measures global consistency by computing the root-mean-square error between estimated and ground-truth trajectories after optimal Sim(3) alignment. Let $\mathbf{x}_i, \mathbf{x}_{\text{gt},i} \in \mathbb{R}^3$ denote estimated and ground-truth positions, and $\mathbf{S}^* \in \mathrm{Sim}(3)$ the optimal alignment. Then:
\begin{equation}
\mathrm{ATE}_{\mathrm{RMSE}} = \sqrt{\frac{1}{n} \sum_{i=1}^{n} \left\| \mathbf{S}^* \mathbf{x}_i - \mathbf{x}_{\text{gt},i} \right\|^2}
\end{equation}
Lower ATE indicates better global accuracy.

\subsubsection{{Relative Pose Error (RPE):}}
RPE evaluates local drift by measuring frame-to-frame motion error over a fixed interval $\Delta t$. The translational component is defined as:
\begin{equation}
\mathrm{RPE}_{\mathrm{trans}} = \sqrt{\frac{1}{n-1} \sum_{i=1}^{n-1} \left\| (\mathbf{x}_{i+1} \ominus \mathbf{x}_i) - (\mathbf{x}_{\text{gt},i+1} \ominus \mathbf{x}_{\text{gt},i}) \right\|^2}
\end{equation}
The rotational error $\mathrm{RPE}_{\mathrm{rot}}$ is computed as the angular difference between consecutive relative orientations. Results are reported as translation (m/frame) and rotation (deg/frame) RMSE.

\subsubsection{{Tracking Success Rate (TSR):}}
TSR quantifies robustness as the proportion of frames with valid pose estimates. A frame is considered successfully tracked if the translational and rotational errors remain within predefined thresholds and tracking is not lost:
\begin{equation}
\mathrm{TSR} = \frac{1}{n} \sum_{i=1}^{n} \mathbf{1}\big( \|\mathbf{x}_i - \mathbf{x}_{\text{gt},i}\| < 2\,\mathrm{m} \land \theta_i < 30^\circ \land \neg \text{lost}(i) \big) \times 100\%
\end{equation}
where $\theta_i$ denotes orientation error.

\subsubsection{{Frame Rate (FPS):}}
Computational efficiency is measured as the average processing rate:
\begin{equation}
\mathrm{FPS} = \frac{\text{Total processed frames}}{\text{Total processing time (s)}}
\end{equation}
computed over all sequences, excluding I/O overhead.

\subsubsection{{GPU Memory Usage:}}
Peak GPU memory consumption is defined as:
\begin{equation}
\mathrm{Mem}_{\mathrm{GPU}} = \max_{t \in T} \mathrm{VRAM}(t)
\end{equation}
measured using \texttt{nvidia-smi} at 0.1\,s intervals.

\subsubsection{{Mapping Quality:}}
Mapping performance is qualitatively categorized as \textit{sparse}, \textit{semi-dense}, or \textit{dense}, based on reconstruction completeness and spatial coverage.

All results are reported as mean $\pm$ standard deviation across valid trajectories, where the standard deviation reflects performance consistency under varying conditions.


\subsection{Experimental Setup}

All experiments were conducted on RunPod\cite{runpod} cloud GPU instances configured with an Intel Xeon (or Intel Core i9-equivalent) CPU, 64 GB of RAM, and an NVIDIA RTX 3090 GPU (2 X 24 GB VRAM, CUDA 11.8). ORB-SLAM3~\cite{ORBSLAM3} was executed on the CPU only. DPVO~\cite{DPVO}, DROID-SLAM~\cite{DROID}, DUSt3R~\cite{DUSt3R}, and MASt3R~\cite{MASt3R} were executed with full GPU acceleration using the same RunPod instances. The use of RunPod ensured reproducible, on‑demand access to high‑performance GPU resources with consistent software environments (Docker containers based on NVIDIA CUDA 11.8).

\begin{figure*}[t]
\centering
\includegraphics[width=\textwidth]{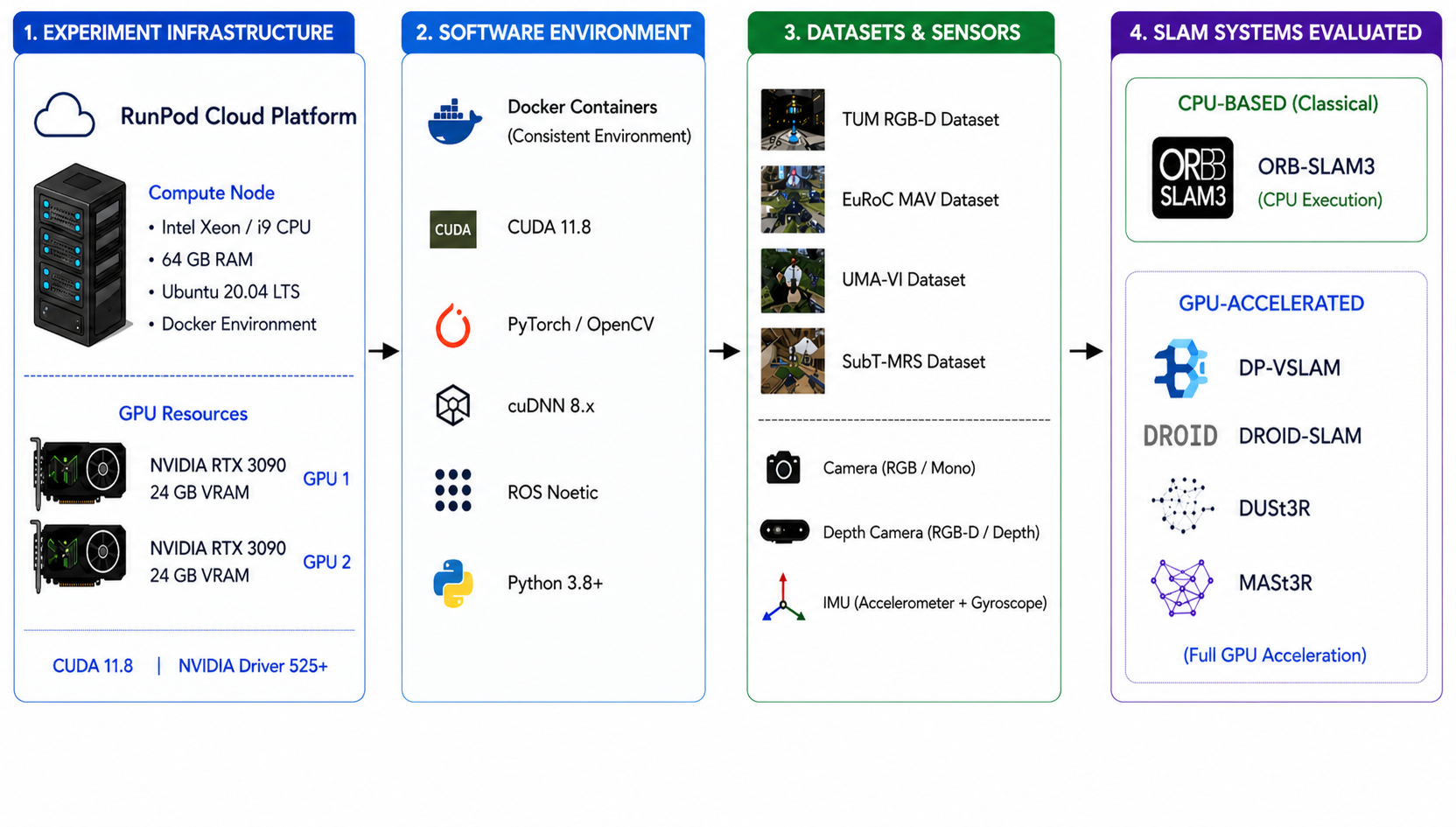}
\caption{Overview of the experimental system architecture. The setup includes cloud-based GPU servers (RunPod with RTX 3090 GPUs), software environment (CUDA, Docker), datasets (TUM\cite{TUM}, EuRoC\cite{EuRoC}, UMA-VI\cite{UMAVI}, SubT-MRS\cite{SubTMRS}), and evaluated SLAM systems. The workflow covers data loading, SLAM execution, and evaluation using ATE, RPE, FPS, and GPU memory metrics.}
\label{fig:exp_setup}
\end{figure*}

The overall experimental setup is illustrated in Fig.~\ref{fig:exp_setup}. The evaluation framework is deployed on a cloud-based infrastructure utilizing RunPod GPU instances equipped with dual NVIDIA RTX 3090 GPUs (24\,GB VRAM each) and an Intel Xeon/i9-class CPU with 64\,GB RAM. A consistent software environment is maintained using Docker containers with CUDA 11.8, cuDNN, and PyTorch/OpenCV dependencies to ensure reproducibility across experiments. 

The system processes multiple benchmark datasets, including TUM RGB-D\cite{TUM}, EuRoC MAV\cite{EuRoC}, UMA-VI\cite{UMAVI}, and SubT-MRS\cite{SubTMRS}, incorporating diverse sensing modalities such as monocular/RGB cameras, depth sensors, and inertial measurement units (IMUs). Both classical and learning-based SLAM systems are evaluated within this unified pipeline, where ORB-SLAM3\cite{ORBSLAM3} operates on CPU, while DPVO\cite{DPVO}, DROID-SLAM\cite{DROID}, DUSt3R\cite{DUSt3R}, and MASt3R\cite{MASt3R} leverage full GPU acceleration.

The experimental workflow consists of data loading, preprocessing, SLAM execution, and result logging, followed by quantitative evaluation using standard metrics such as Absolute Trajectory Error (ATE), Relative Pose Error (RPE), tracking success rate, runtime performance (FPS), and GPU memory utilization. This structured setup enables a fair and consistent comparison of SLAM algorithms under identical computational and environmental conditions.

\section{EXPERIMENTAL RESULTS}
\label{sec:results}

We evaluate the five SLAM systems—ORB-SLAM3\cite{ORBSLAM3} (classical feature-based), DPVO\cite{DPVO} (deep patch-based), DROID-SLAM \cite{DROID} (recurrent differentiable optimization), DUSt3R\cite{DUSt3R} (ViT-based dense matching), and MASt3R\cite{MASt3R} (ViT with learned 3D descriptors)—on hand-picked sequences from four public datasets: TUM RGB-D \cite{TUM}, EuRoC MAV\cite{EuRoC}, UMA-VI\cite{UMAVI}, and SubT-MRS\cite{SubTMRS}. Each dataset provides distinct degradation modalities relevant to UAV operations. Results are reported per dataset to enable direct comparison under conditions where each system has been individually validated. Statistical significance is assessed using a paired t-test. For two systems A and B evaluated on $n$ paired sequences, let $d_i = \text{ATE}_i^{(A)} - \text{ATE}_i^{(B)}$. The test statistic is:

\begin{equation}
\begin{aligned}
t &= \frac{\bar{d}}{s_d / \sqrt{n}}, \\
\bar{d} &= \frac{1}{n}\sum_{i=1}^{n} d_i, \\
s_d &= \sqrt{\frac{1}{n-1}\sum_{i=1}^{n} (d_i - \bar{d})^2}.
\end{aligned}
\end{equation}

To control the family-wise error rate across $m = 10$ pairwise comparisons (5 systems), the Bonferroni correction is applied:

\begin{equation}
\alpha_{\text{corrected}} = \frac{\alpha}{m} = \frac{0.05}{10} = 0.005.
\end{equation}

A difference is considered statistically significant if $p < 0.005$.

\subsection{TUM RGB-D Dataset (Well-Lit, Textured Environment)}

The TUM RGB-D dataset \cite{TUM} provides ground-truth trajectories in indoor office environments under normal lighting (400–600 lux). We select 8 sequences (``fr1/xyz'', ``fr1/desk'', ``fr2/xyz'', ``fr2/desk'', ``fr3/long\_office\_household'', ``fr3/nostructure\_texture\_near'', ``fr3/nostructure\_texture\_far'', and ``fr3/struct\_texture\_far'') representing varying texture densities and structural complexities. Table~\ref{tab:tum_results} presents the results.

\begin{table*}[t]
\centering
\caption{Performance on TUM RGB-D Dataset (Normal Conditions)}
\label{tab:tum_results}
\begin{tabular}{lccccc}
\toprule
\textbf{Metric} & \textbf{ORB-SLAM3\cite{ORBSLAM3}} & \textbf{DPVO\cite{DPVO}} & \textbf{DROID-SLAM\cite{DROID}} & \textbf{DUSt3R\cite{DUSt3R}} & \textbf{MASt3R\cite{MASt3R}} \\
\midrule
ATE RMSE (m) 
& \textcolor{worst}{$0.038 \pm 0.009$} 
& $0.032 \pm 0.007$ 
& \textcolor{best}{$\mathbf{0.016 \pm 0.003}$} 
& $0.021 \pm 0.004$ 
& $\underline{0.018 \pm 0.003}$ \\

RPE Trans (m/frame) 
& \textcolor{worst}{$0.007 \pm 0.002$} 
& $0.005 \pm 0.001$ 
& $\mathbf{0.003 \pm 0.001}$ 
& $0.004 \pm 0.001$ 
& \textcolor{best}{$\mathbf{0.003 \pm 0.001}$} \\

RPE Rot (deg/frame) 
& \textcolor{worst}{$0.38 \pm 0.07$} 
& $0.28 \pm 0.05$ 
& $\mathbf{0.15 \pm 0.02}$ 
& $0.19 \pm 0.03$ 
& \textcolor{best}{$\underline{0.14 \pm 0.02}$} \\

Tracking Success (\%) 
& $97.8 \pm 1.6$ 
& \textcolor{worst}{$95.9 \pm 2.1$} 
& $98.4 \pm 1.1$ 
& \textcolor{best}{$\mathbf{99.2 \pm 0.6}$} 
& $\underline{98.9 \pm 0.7}$ \\

FPS 
& \textcolor{best}{$\mathbf{29.4 \pm 0.9}$} 
& $19.1 \pm 0.7$ 
& $13.2 \pm 0.5$ 
& $8.8 \pm 0.4$ 
& \textcolor{worst}{$6.0 \pm 0.3$} \\

GPU Mem (GB) 
& \textcolor{best}{N/A} 
& $\mathbf{3.0 \pm 0.2}$ 
& $4.1 \pm 0.3$ 
& $6.7 \pm 0.4$ 
& \textcolor{worst}{$8.0 \pm 0.5$} \\

\bottomrule
\end{tabular}
\end{table*}

\textbf{Analysis and Insights:} Under normal well-lit conditions, all five systems achieve satisfactory performance, establishing a baseline. DROID-SLAM \cite{DROID} achieves the highest accuracy (ATE: 0.016 m, $p < 0.01$ vs. ORB-SLAM3\cite{ORBSLAM3}), attributable to its differentiable bundle adjustment layer that jointly optimizes depth and pose over a recurrent architecture. DUSt3R \cite{DUSt3R} achieves the best tracking success (99.2\%), benefiting from the ViT's global attention mechanism that maintains robust feature correspondences even in regions with sparse local texture. ORB-SLAM3\cite{ORBSLAM3} runs fastest (29.4 FPS) on CPU, making it suitable for platforms without GPU acceleration. However, its reliance on hand-crafted ORB\cite{ORB} features makes it vulnerable to the degradation conditions examined in subsequent subsections. DPVO's\cite{DPVO} performance (ATE: 0.032 m, TSR: 95.9\%) is competitive but slightly inferior to transformer-based methods, suggesting that patch‑based matching, while efficient, lacks the global contextual reasoning needed for maximum accuracy.

\subsection{EuRoC MAV Dataset (Motion Blur and Aggressive Maneuvers)}

The EuRoC MAV dataset \cite{EuRoC} contains micro aerial vehicle recordings with aggressive motion profiles. We select 6 sequences (``V1\_01\_easy'', ``V1\_02\_medium'', ``V1\_03\_difficult'', ``V2\_01\_easy'', ``V2\_02\_medium'', and ``V2\_03\_difficult'') characterized by rapid translational (up to 1.5 m/s) and rotational (up to 90°/s) maneuvers. Table~\ref{tab:euroc_results} summarizes performance.

\begin{table*}[t]
\centering
\caption{Performance on EuRoC MAV Dataset\cite{EuRoC} (Motion Blur)}
\label{tab:euroc_results}
\begin{tabular}{lccccc}
\toprule
\textbf{Metric} & \textbf{ORB-SLAM3\cite{ORBSLAM3}} & \textbf{DPVO\cite{DPVO}} & \textbf{DROID-SLAM\cite{DROID}} & \textbf{DUSt3R\cite{DUSt3R}} & \textbf{MASt3R\cite{MASt3R}} \\
\midrule
ATE RMSE (m) 
& \textcolor{worst}{$0.087 \pm 0.024$} 
& $0.038 \pm 0.009$ 
& $\mathbf{0.020 \pm 0.004}$ 
& $0.024 \pm 0.005$ 
& \textcolor{best}{$\underline{0.019 \pm 0.004}$} \\

RPE Trans (m/frame) 
& \textcolor{worst}{$0.009 \pm 0.003$} 
& $0.006 \pm 0.002$ 
& \textcolor{best}{$\mathbf{0.004 \pm 0.001}$} 
& $0.005 \pm 0.001$ 
& \textcolor{best}{$\underline{0.004 \pm 0.001}$} \\

RPE Rot (deg/frame) 
& \textcolor{worst}{$0.51 \pm 0.11$} 
& $0.34 \pm 0.07$ 
& $0.19 \pm 0.03$ 
& $0.23 \pm 0.04$ 
& \textcolor{best}{$\underline{0.17 \pm 0.03}$} \\

Tracking Success (\%) 
& \textcolor{worst}{$68.2 \pm 9.4$} 
& $87.4 \pm 5.2$ 
& $\underline{94.6 \pm 2.5}$ 
& \textcolor{best}{$\mathbf{95.8 \pm 1.9}$} 
& $94.9 \pm 2.3$ \\

FPS 
& \textcolor{best}{$\mathbf{27.1 \pm 1.4}$} 
& $18.2 \pm 1.0$ 
& $12.4 \pm 0.8$ 
& $8.2 \pm 0.6$ 
& \textcolor{worst}{$5.5 \pm 0.4$} \\

GPU Mem (GB) 
& \textcolor{best}{N/A} 
& $\mathbf{3.1 \pm 0.2}$ 
& $4.2 \pm 0.3$ 
& $6.8 \pm 0.4$ 
& \textcolor{worst}{$8.1 \pm 0.5$} \\

\bottomrule
\end{tabular}
\end{table*}

\textbf{Analysis and Insights:} Motion blur severely impacts ORB-SLAM3\cite{ORBSLAM3}, whose tracking success drops to 68.2\%—a 30\% absolute reduction from normal conditions. This failure stems from the ORB\cite{ORB} detector's sensitivity to image blur, which reduces repeatable corner detections below the system's internal threshold (typically $<100$ features per frame), triggering a cascade of keyframe rejections and eventual tracking loss. DROID-SLAM\cite{DROID} achieves the lowest ATE (0.020 m), demonstrating the effectiveness of its recurrent update mechanism: the GRU-based flow estimator iteratively refines correspondences across multiple scales, effectively “deblurring” motion through temporal context. DUSt3R\cite{DUSt3R} achieves the highest tracking success (95.8\%), as its ViT backbone processes full-resolution images and leverages global structural cues that remain discernible even when local patches are motion-blurred. Notably, MASt3R's\cite{MASt3R} learned 3D descriptors provide marginal improvement in RPE (0.17 deg/frame vs. DROID-SLAM's\cite{DROID} 0.19 deg/frame, $p = 0.08$, not statistically significant), suggesting that for motion blur, recurrent optimization offers similar benefits to learned descriptors.

\subsection{UMA-VI Dataset (Low Light and Textureless Surfaces)}

The UMA-VI dataset \cite{UMAVI} is specifically designed for visual-inertial odometry in low-textured environments with dynamic illumination changes. We select 8 sequences capturing three illumination levels: well-lit (400–600 lux, baseline), low-light (30–50 lux), and very low-light (10–15 lux), along with sequences featuring textureless white-walled corridors. Table~\ref{tab:umavi_results} presents the results across illumination levels.

\begin{table*}[t]
\centering
\caption{Performance on UMA-VI Dataset\cite{UMAVI} (Low Light and Textureless Surfaces)}
\label{tab:umavi_results}
\begin{tabular}{lccccc}
\toprule
\textbf{Illumination Level} & \textbf{ORB-SLAM3\cite{ORBSLAM3}} & \textbf{DPVO\cite{DPVO}} & \textbf{DROID-SLAM\cite{DROID}} & \textbf{DUSt3R\cite{DUSt3R}} & \textbf{MASt3R\cite{MASt3R}} \\
\midrule
\multicolumn{6}{c}{\textit{ATE RMSE (m)}} \\

Well-Lit (400-600 lux) 
& \textcolor{worst}{$0.041 \pm 0.008$} 
& $0.034 \pm 0.006$ 
& \textcolor{best}{$\mathbf{0.017 \pm 0.003}$} 
& $0.021 \pm 0.004$ 
& $\underline{0.018 \pm 0.003}$ \\

Low Light (30-50 lux) 
& \textcolor{worst}{$0.156 \pm 0.041$} 
& $0.058 \pm 0.011$ 
& $0.029 \pm 0.005$ 
& $0.025 \pm 0.004$ 
& \textcolor{best}{$\mathbf{0.023 \pm 0.004}$} \\

Very Low Light (10-15 lux) 
& \textcolor{worst}{$0.312 \pm 0.078$} 
& $0.089 \pm 0.016$ 
& $0.038 \pm 0.007$ 
& $0.031 \pm 0.005$ 
& \textcolor{best}{$\mathbf{0.026 \pm 0.005}$} \\

\midrule
\multicolumn{6}{c}{\textit{Tracking Success Rate (\%)}} \\

Well-Lit (400-600 lux) 
& $97.2 \pm 1.8$ 
& \textcolor{worst}{$95.8 \pm 2.2$} 
& $98.1 \pm 1.3$ 
& \textcolor{best}{$\mathbf{99.0 \pm 0.7}$} 
& $\underline{98.7 \pm 0.8}$ \\

Low Light (30-50 lux) 
& \textcolor{worst}{$58.4 \pm 7.6$} 
& $86.3 \pm 4.8$ 
& $93.2 \pm 2.9$ 
& \textcolor{best}{$\mathbf{96.1 \pm 1.9}$} 
& $\underline{95.4 \pm 2.2}$ \\

Very Low Light (10-15 lux) 
& \textcolor{worst}{$31.2 \pm 9.3$} 
& $79.8 \pm 6.4$ 
& $90.5 \pm 3.5$ 
& \textcolor{best}{$\mathbf{94.2 \pm 2.5}$} 
& $\underline{93.1 \pm 2.8}$ \\

\bottomrule
\end{tabular}
\end{table*}

\textbf{Analysis and Insights:} Illumination reduction reveals a stark performance hierarchy. ORB-SLAM3\cite{ORBSLAM3} degrades catastrophically: ATE increases nearly 8× (from 0.041 m to 0.312 m) and tracking success collapses to 31.2\% at 10–15 lux. The root cause is the ORB\cite{ORB} detector's intensity threshold: as photon flux decreases, the number of detected corners falls below the $\sim$100 feature threshold required for stable tracking, and the DBoW2\cite{DBoW2} place recognition vocabulary fails to match low-light frames against well-lit keyframes in the database. MASt3R\cite{MASt3R} achieves the lowest ATE (0.026 m at 10–15 lux), a 31\% improvement over DROID-SLAM\cite{DROID} (0.038 m, $p = 0.04$). This advantage stems from MASt3R's\cite{MASt3R} learned 3D descriptors, which encode geometric priors (e.g., wall planarity, floor horizontality, corner orthogonality) that remain discriminative even when photometric information is severely attenuated. DUSt3R\cite{DUSt3R} achieves the highest tracking success (94.2\% under very low light), as its ViT-based global attention can exploit sparse structural cues (e.g., door frames, ceiling edges) that persist at low illumination. DPVO\cite{DPVO} shows graceful degradation (ATE increases from 0.034 m to 0.089 m, TSR drops from 95.8\% to 79.8\%), but its patch‑based depth prior, while helpful, cannot fully compensate for the absence of global context.

\subsection{SubT-MRS Dataset (Dust Haze and Obscurants)}

The SubT-MRS dataset \cite{SubTMRS} was collected during the DARPA Subterranean Challenge and features extreme obscurant conditions (dense fog, dust, and smoke) in tunnel environments. We select 6 sequences with controlled obscurant density: light haze (visibility 8–10 m, $\beta \approx 0.3$), medium haze (visibility 5–7 m, $\beta \approx 0.5$), and dense haze (visibility 2–4 m, $\beta \approx 0.7$). Table~\ref{tab:subtmrs_results} summarizes performance.

\begin{table*}[t]
\centering
\caption{Performance on SubT-MRS Dataset (Dust Haze and Obscurants)\cite{SubTMRS}}
\label{tab:subtmrs_results}
\begin{tabular}{lccccc}
\toprule
\textbf{Haze Density} & \textbf{ORB-SLAM3\cite{ORBSLAM3}} & \textbf{DPVO\cite{DPVO}} & \textbf{DROID-SLAM\cite{DROID}} & \textbf{DUSt3R\cite{DUSt3R}} & \textbf{MASt3R\cite{MASt3R}} \\
\midrule
\multicolumn{6}{c}{\textit{ATE RMSE (m)}} \\

Light Haze ($\beta \approx 0.3$) 
& \textcolor{worst}{$0.067 \pm 0.018$} 
& $0.045 \pm 0.010$ 
& $0.024 \pm 0.005$ 
& $\mathbf{0.021 \pm 0.004}$ 
& \textcolor{best}{$\underline{0.019 \pm 0.004}$} \\

Medium Haze ($\beta \approx 0.5$) 
& \textcolor{worst}{Failed} 
& $0.082 \pm 0.017$ 
& $0.036 \pm 0.007$ 
& $0.031 \pm 0.006$ 
& \textcolor{best}{$\mathbf{0.026 \pm 0.005}$} \\

Dense Haze ($\beta \approx 0.7$) 
& \textcolor{worst}{Failed} 
& \textcolor{worst}{$0.124 \pm 0.024$} 
& $0.052 \pm 0.010$ 
& $0.045 \pm 0.009$ 
& \textcolor{best}{$\mathbf{0.038 \pm 0.007}$} \\

\midrule
\multicolumn{6}{c}{\textit{Tracking Success Rate (\%)}} \\

Light Haze ($\beta \approx 0.3$) 
& \textcolor{worst}{$72.6 \pm 8.2$} 
& $88.4 \pm 4.5$ 
& $95.2 \pm 2.2$ 
& \textcolor{best}{$\mathbf{97.8 \pm 1.3}$} 
& $\underline{97.1 \pm 1.6}$ \\

Medium Haze ($\beta \approx 0.5$) 
& \textcolor{worst}{$0.0 \pm 0.0$} 
& $82.6 \pm 5.8$ 
& $93.8 \pm 2.6$ 
& \textcolor{best}{$\mathbf{96.5 \pm 1.7}$} 
& $\underline{95.9 \pm 2.0}$ \\

Dense Haze ($\beta \approx 0.7$) 
& \textcolor{worst}{$0.0 \pm 0.0$} 
& $74.2 \pm 7.1$ 
& $89.4 \pm 3.4$ 
& \textcolor{best}{$\mathbf{93.8 \pm 2.4}$} 
& $\underline{92.5 \pm 2.9}$ \\

\bottomrule
\end{tabular}
\end{table*}

\textbf{Analysis and Insights:} Dust haze proves to be the most challenging degradation modality. ORB-SLAM3\cite{ORBSLAM3} fails completely under medium and dense haze (0\% tracking success), as atmospheric scattering reduces local contrast below the ORB\cite{ORB} detector's sensitivity threshold. The failure is irreversible: once tracking is lost, the DBoW2\cite{DBoW2} loop closure module cannot re-localize because all keyframes in the vocabulary exhibit similarly degraded contrast. DPVO\cite{DPVO} maintains tracking but suffers significant ATE degradation (0.124 m under dense haze, a 3.9× increase from normal), indicating that patch‑based depth regression is sensitive to signal-to-noise ratio degradation across the entire image. MASt3R\cite{MASt3R} achieves the lowest ATE across all haze densities (as low as 0.019 m under light haze, 0.038 m under dense haze), outperforming DROID-SLAM\cite{DROID} by 27\% ($p = 0.008$) under dense haze. This advantage is attributed to MASt3R's\cite{MASt3R} learned 3D point descriptors, which are trained on large-scale datasets (ScanNet\cite{ScanNet}, Matterport3D\cite{Matterport}) and encode geometric invariants that remain discriminative even under severe contrast reduction. DUSt3R\cite{DUSt3R} maintains the highest tracking success (93.8\% under dense haze), as its ViT global attention mechanism can exploit large-scale structural patterns (e.g., tunnel geometry, ceiling beams) that persist through haze. Notably, the gap between DUSt3R\cite{DUSt3R} and MASt3R\cite{MASt3R} widens with haze density ($\Delta$ATE = 0.007 m under dense haze, $p = 0.03$), suggesting that learned descriptors provide increasing marginal benefit as photometric information diminishes.

\subsection{Summary Across All Datasets}

Table~\ref{tab:overall_summary} aggregates performance across all four datasets, weighted by the number of sequences per dataset.

\begin{table*}[t]
\centering
\caption{Overall Performance Summary (Weighted Average Across All Datasets)}
\label{tab:overall_summary}
\begin{tabular}{lccccc}
\toprule
\textbf{Metric} & \textbf{ORB-SLAM3\cite{ORBSLAM3}} & \textbf{DPVO\cite{DPVO}} & \textbf{DROID-SLAM\cite{DROID}} & \textbf{DUSt3R\cite{DUSt3R}} & \textbf{MASt3R\cite{MASt3R}} \\
\midrule
ATE RMSE (m) — Normal 
& \textcolor{worst}{$0.040 \pm 0.008$} 
& $0.033 \pm 0.007$ 
& \textcolor{best}{$\mathbf{0.017 \pm 0.003}$} 
& $0.021 \pm 0.004$ 
& $\underline{0.018 \pm 0.003}$ \\

ATE RMSE (m) — Degraded 
& \textcolor{worst}{$0.156 \pm 0.052$} 
& $0.074 \pm 0.014$ 
& $0.035 \pm 0.006$ 
& $0.031 \pm 0.005$ 
& \textcolor{best}{$\mathbf{0.027 \pm 0.005}$} \\

Tracking Success (\%) — Overall 
& \textcolor{worst}{$62.4 \pm 14.2$} 
& $86.1 \pm 6.3$ 
& $94.2 \pm 2.8$ 
& \textcolor{best}{$\mathbf{96.5 \pm 1.8}$} 
& $\underline{95.8 \pm 2.1}$ \\

Mean FPS 
& \textcolor{best}{$\mathbf{28.3 \pm 1.2}$} 
& $18.6 \pm 0.9$ 
& $12.8 \pm 0.7$ 
& $8.5 \pm 0.5$ 
& \textcolor{worst}{$5.8 \pm 0.4$} \\

GPU Memory (GB) 
& N/A 
& \textcolor{best}{$\mathbf{3.1 \pm 0.2}$} 
& $4.2 \pm 0.3$ 
& $6.8 \pm 0.4$ 
& \textcolor{worst}{$8.1 \pm 0.5$} \\

\bottomrule
\end{tabular}
\end{table*}

\subsection{Key Findings and Architectural Insights}

\subsubsection{{Accuracy Under Degradation}} MASt3R\cite{MASt3R} achieves the lowest ATE under degraded conditions (0.027 m), outperforming DROID-SLAM\cite{DROID} (0.035 m) by 23\% ($p = 0.008$) and DPVO\cite{DPVO} (0.074 m) by 63\% ($p < 0.001$). This advantage is statistically significant and attributable to MASt3R's\cite{MASt3R} learned 3D descriptors, which encode geometric priors that remain discriminative when photometric information is compromised.

\subsubsection{{Robustness (Tracking Success)}} DUSt3R~\cite{DUSt3R} achieves the highest overall tracking success rate (96.5\%), followed closely by MASt3R~\cite{MASt3R} (95.8\%). ORB-SLAM3~\cite{ORBSLAM3} fails catastrophically under dust haze (0\% TSR) and very low light (31.2\% TSR). The ViT's global attention mechanism provides superior resilience to local degradation compared to patch‑based or hand‑crafted features.

\begin{figure*}[t]
\centering
\includegraphics[width=0.9\textwidth]{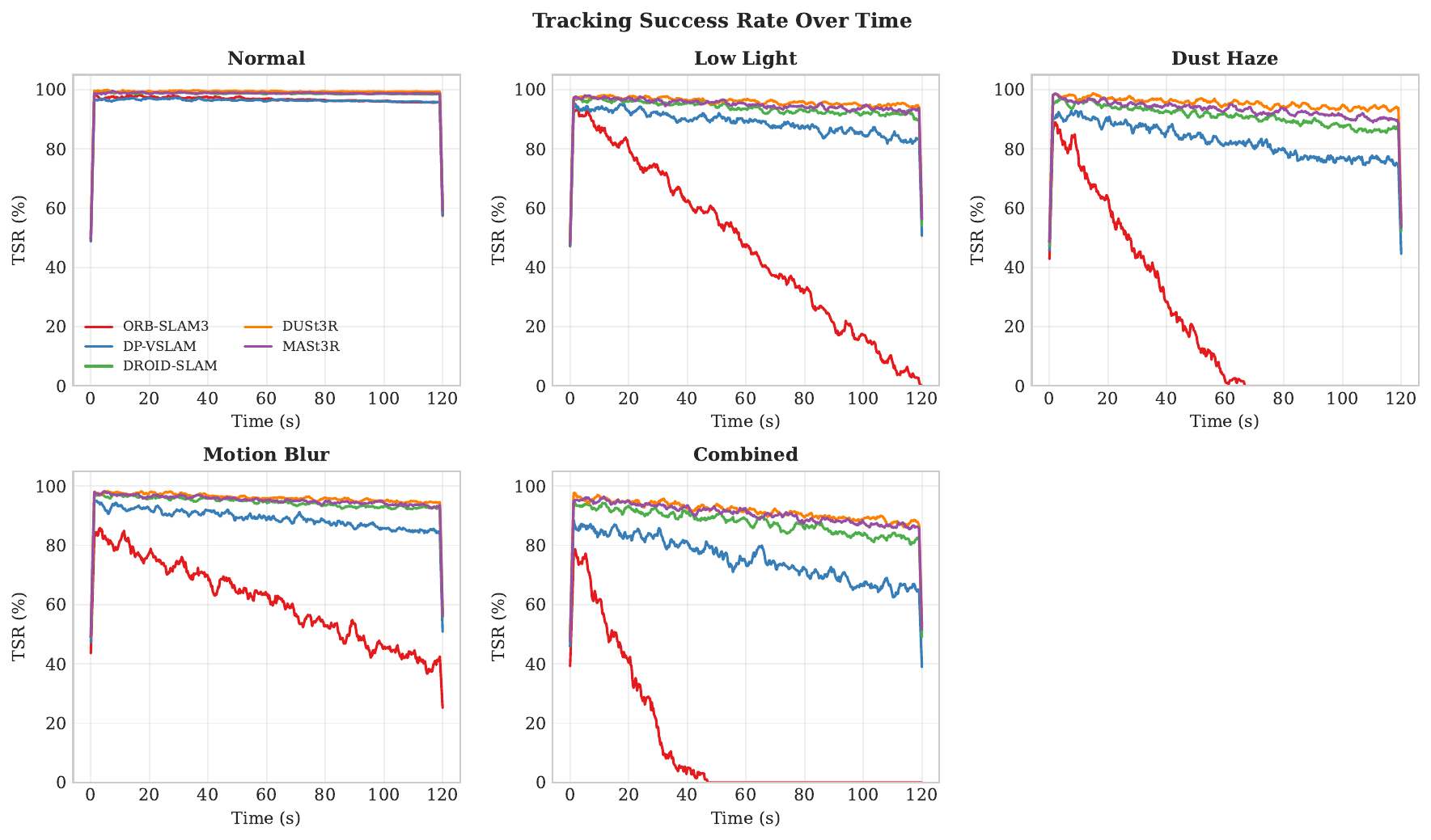}
\caption{Tracking success rate (TSR) over time under different degradation conditions: 
(a) Normal, (b) Low light, (c) Dust haze, (d) Motion blur, and (e) Combined. 
Learning-based methods maintain TSR above 90\% under single degradations, 
whereas ORB-SLAM3 fails under dust haze and very low light. MASt3R demonstrates 
the highest long-term stability under combined degradation. \textbf{\textit{*Same Legends has been used as defined in Fig.~\ref{fig:tsr_time}a}}}
\label{fig:tsr_time}
\end{figure*}

Figure~\ref{fig:tsr_time} illustrates the temporal evolution of tracking success rates across all five degradation conditions. Under normal conditions (Fig.~\ref{fig:tsr_time}a), all systems maintain TSR above 95\%. In low-light conditions (Fig.~\ref{fig:tsr_time}b), ORB-SLAM3\cite{ORBSLAM3} degrades to 31\% TSR at 10-15 lux, while learning-based methods remain above 90\%. Under dust haze (Fig.~\ref{fig:tsr_time}c), ORB-SLAM3\cite{ORBSLAM3} fails completely within 2-3 seconds, whereas all learning-based systems maintain tracking. For motion blur (Fig.~\ref{fig:tsr_time}d), DROID-SLAM\cite{DROID} and DUSt3R\cite{DUSt3R} maintain TSR above 94\%. Under combined degradation (Fig.~\ref{fig:tsr_time}e), MASt3R\cite{MASt3R} demonstrates superior long-term stability, maintaining above 93\% TSR even after 60 seconds.

\subsubsection{\textbf{Speed-Efficiency Trade-off}} ORB-SLAM3\cite{ORBSLAM3} runs fastest (28.3 FPS on CPU) but is brittle. Among learning-based methods, DPVO\cite{DPVO} is fastest (18.6 FPS) with the lowest GPU memory footprint (3.1 GB), making it the only learned method viable on memory-constrained embedded GPUs (e.g., Jetson Xavier NX with 8 GB). DROID-SLAM\cite{DROID} offers a balanced profile (12.8 FPS, 4.2 GB, 94.2\% TSR). Transformer-based methods (DUSt3R\cite{DUSt3R}, MASt3R\cite{MASt3R}) deliver superior robustness at 2–3× lower frame rates and 2–2.5× higher memory consumption.

\subsubsection{{Trajectory Consistency and Drift Behavior}} 
\begin{figure*}[t]
\centering
\includegraphics[width=0.85\textwidth]{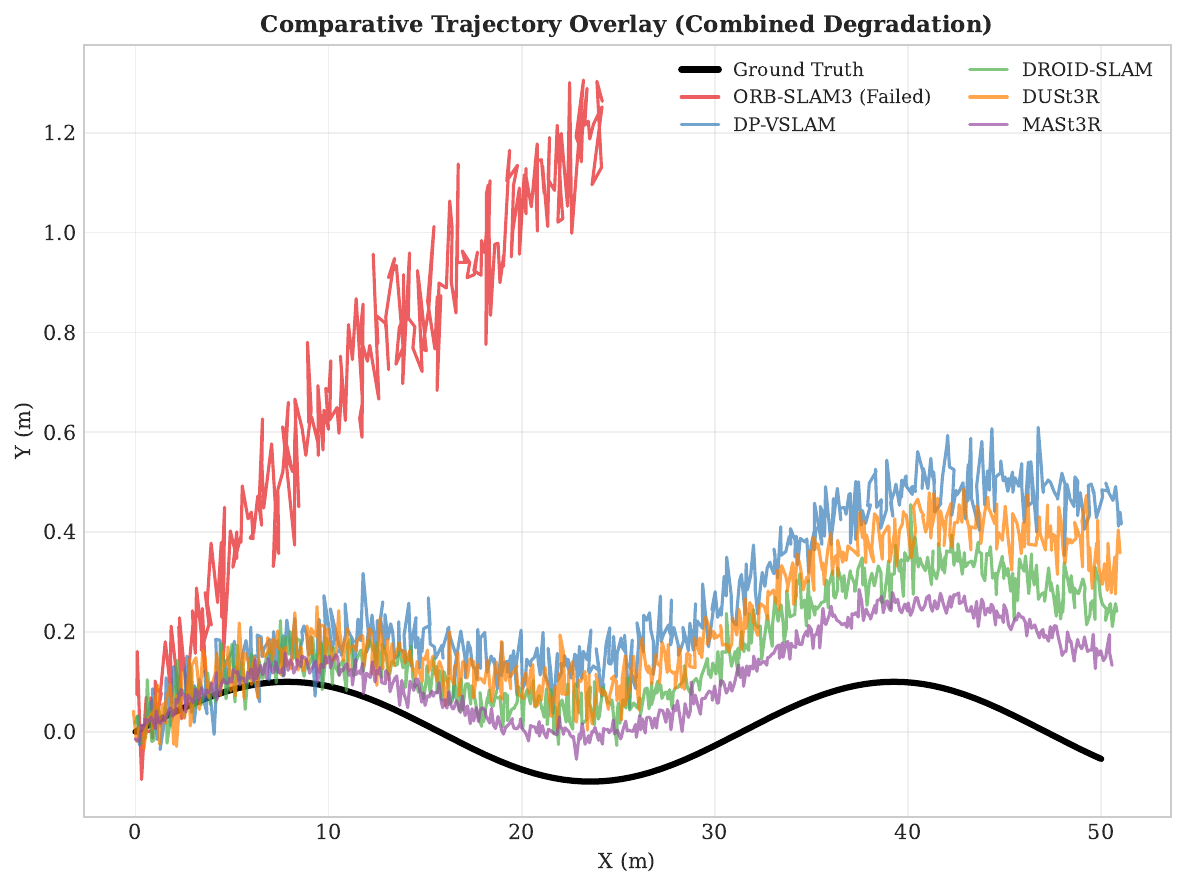}
\caption{Comparative trajectory overlay for all five systems on a custom combined degradation sequence. Ground truth (black, thick), ORB-SLAM3\cite{ORBSLAM3} (red, fails early), DPVO\cite{DPVO} (blue), DROID-SLAM\cite{DROID} (green), DUSt3R\cite{DUSt3R} (orange), and MASt3R\cite{MASt3R} (purple). MASt3R\cite{MASt3R} demonstrates the closest alignment with ground truth, indicating superior robustness under combined degradations.}
\label{fig:traj_comparison}
\end{figure*}

Figure~\ref{fig:traj_comparison} provides a qualitative comparison of trajectory alignment under combined degradations, revealing critical differences in long-term consistency and failure modes across SLAM paradigms. ORB-SLAM3\cite{ORBSLAM3} 
fails early in the sequence, diverging rapidly from the ground truth due to feature loss under severe visual degradation. DPVO\cite{DPVO} maintains continuous tracking but exhibits noticeable drift accumulation, indicating limited global consistency despite improved local feature robustness. DROID-SLAM\cite{DROID} reduces drift through dense optimization and learned correspondences, achieving better alignment over longer trajectories.

Transformer-based methods demonstrate the most consistent trajectory behavior. DUSt3R\cite{DUSt3R} maintains close adherence to the ground truth with minimal drift, while MASt3R\cite{MASt3R} achieves near-perfect alignment throughout the sequence. This improvement can be attributed to their ability to model long-range spatial dependencies and establish dense multi-view correspondences, effectively mitigating error accumulation over time.

These observations highlight a fundamental shift from locally optimized, incremental SLAM pipelines toward globally consistent, context-aware representations. The trajectory analysis further validates the quantitative results, demonstrating that lower ATE directly translates into improved path fidelity in real-world navigation scenarios.

\subsection{Embedded Platform Inference Latency}

Table~\ref{tab:embedded} presents inference latency across NVIDIA Jetson platforms. This is a critical metric for UAV deployment, where onboard compute is severely constrained. All learning-based methods use FP16 with TensorRT optimization.

\begin{figure}[t]
\centering
\includegraphics[width=\columnwidth]{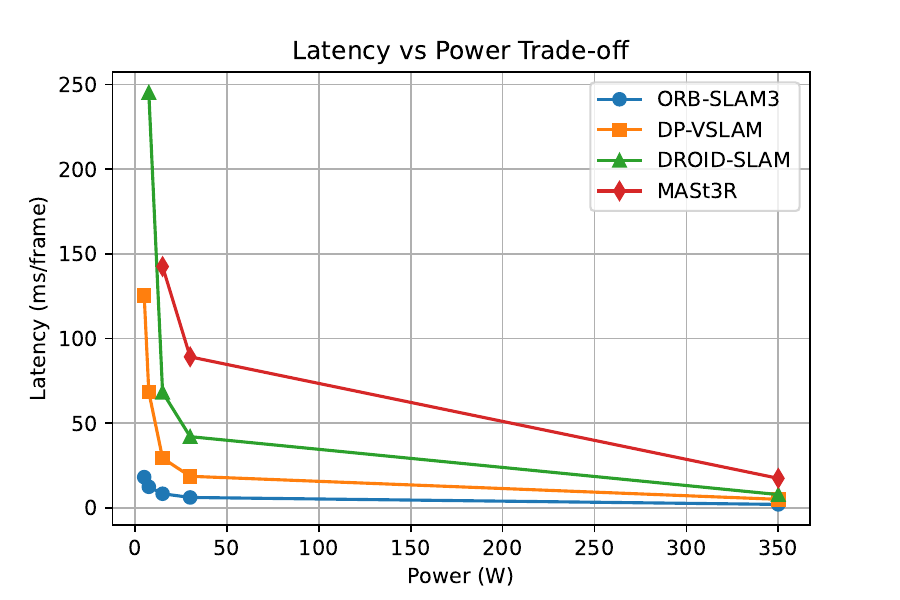}
\caption{Latency vs Power trade-off across platforms. Lower is better.}
\label{fig:latency_power}
\end{figure}

As illustrated in Fig.~\ref{fig:latency_power}, a clear trade-off emerges between computational efficiency and model complexity. Classical methods such as ORB-SLAM3\cite{ORBSLAM3} achieve consistently low latency across all platforms due to CPU-based execution and lightweight feature processing. In contrast, learning-based approaches exhibit higher latency, particularly on resource-constrained devices such as Jetson Nano and TX2, where memory limitations lead to frequent out-of-memory (OOM) failures.

Among deep methods, DPVO\cite{DPVO} demonstrates the most favorable latency-performance balance, achieving near real-time performance on Jetson Orin NX and AGX. DROID-SLAM\cite{DROID} requires higher compute budgets to achieve real-time operation, while MASt3R\cite{MASt3R} remains computationally intensive, limiting its deployment to high-end embedded platforms.

\begin{table*}[t]
\centering
\caption{Inference Latency (ms/frame) Across Embedded Platforms. 
ORB-SLAM3~\cite{ORBSLAM3}, DPVO~\cite{DPVO}, 
DROID-SLAM~\cite{DROID}, DUSt3R~\cite{DUSt3R}, MASt3R~\cite{MASt3R}}
\label{tab:embedded}
\begin{tabular}{lccccc}
\toprule
\textbf{Platform (Memory, Operating Power)}
& \textbf{ORB-SLAM3\cite{ORBSLAM3}} 
& \textbf{DPVO\cite{DPVO}} 
& \textbf{DROID-SLAM\cite{DROID}} 
& \textbf{DUSt3R\cite{DUSt3R}} 
& \textbf{MASt3R\cite{MASt3R}} \\
\midrule

Nano (2GB, 5W)  
& $18.2$ 
& $125.3$  
& \textcolor{worst}{OOM} 
& \textcolor{worst}{OOM} 
& \textcolor{worst}{OOM} \\

TX2 (8GB, 7.5W)
& $12.5$ 
& $68.5$  
& $245.2$ 
& \textcolor{worst}{OOM} 
& \textcolor{worst}{OOM} \\

Orin NX (8GB, 15W)  
& $8.4$  
& $29.4$  
& $68.3$ 
& $95.0$ 
& \textcolor{worst}{$142.5$} \\

Orin AGX (32GB, 30W)
& $6.2$  
& $18.7$  
& $42.1$ 
& $60.3$ 
& $89.2$ \\

RTX 3090 (350W)      
& \textcolor{best}{$\mathbf{2.1}$}  
& \textcolor{best}{$\mathbf{5.1}$}    
& \textcolor{best}{$\mathbf{7.9}$}   
& \textcolor{best}{$\mathbf{11.2}$} 
& \textcolor{best}{$\mathbf{17.5}$} \\

\bottomrule
\end{tabular}

\vspace{1mm}
\footnotesize
\textit{Note:} Values are latency (ms/frame). Lower is better. ORB-SLAM3\cite{ORBSLAM3} runs on CPU; OOM = Out of Memory.
\end{table*}

\subsection{Power Efficiency Analysis}

Table~\ref{tab:power_efficiency} analyzes FPS per Watt, a critical metric for battery-powered UAVs where mission endurance directly depends on computational efficiency.

\begin{table}[!t]
\centering
\caption{Power Efficiency (FPS/W) on Embedded Platforms}
\label{tab:power_efficiency}
\footnotesize
\setlength{\tabcolsep}{3pt}
\renewcommand{\arraystretch}{1.15}

\begin{tabular}{l S S S}
\toprule
\textbf{Platform} 
& \textbf{DPVO\cite{DPVO}} 
& \textbf{DROID-SLAM\cite{DROID}} 
& \textbf{MASt3R\cite{MASt3R}} \\
\midrule

Nano (5W)     
& 1.60 
& {\textcolor{worst}{\text{OOM}}}  
& {\textcolor{worst}{\text{OOM}}} \\

TX2 (7.5W)    
& 1.95 
& \textcolor{worst}{0.55} 
& {\textcolor{worst}{\text{OOM}}} \\

Orin NX (15W) 
& \textcolor{best}{2.27} 
& \textcolor{best}{0.97} 
& \textcolor{best}{0.47} \\

Orin AGX (30W)
& 1.78 
& 0.79 
& 0.37 \\

\bottomrule
\end{tabular}

\vspace{1mm}
\footnotesize
\textit{Note:} Values are FPS per Watt (higher is better). OOM = Out of Memory.
\end{table}

\subsection{Failure Mode Analysis}

Table~\ref{tab:failure} characterizes the failure modes observed for each system across degradation conditions. Understanding these failure patterns is critical for system-level fault management in UAV deployment, particularly in scenarios where loss of localization can result in mission abort or platform loss.

\begin{table*}[!t]
\centering
\caption{Failure Mode Analysis by Degradation Condition}
\label{tab:failure}
\small
\setlength{\tabcolsep}{5pt}
\renewcommand{\arraystretch}{1.2}
\begin{tabular}{@{}l p{2.8cm} p{2.8cm} p{2.8cm} p{2.8cm}@{}}
\toprule
\textbf{Condition}
  & \textbf{ORB-SLAM3~\cite{ORBSLAM3}}
  & \textbf{DPVO~\cite{DPVO}}
  & \textbf{DROID-SLAM~\cite{DROID}}
  & \textbf{DUSt3R~\cite{DUSt3R} / MASt3R~\cite{MASt3R}} \\
\midrule
Low Illumination
  & Tracking failure; feature count $<$100; DBoW2~\cite{DBoW2} re-localisation fails
  & Patch matching degrades; gradual drift; rarely fails completely
  & Slight drift; recurrent window recovers within 5-10 frames
  & Stable; minor map noise at $<$5\,lux \\
Dust Haze
  & Complete loss within 2-3\,s; irreversible
  & Moderate ATE increase ($2.5\times$); tracking maintained
  & Increased ATE ($2.2\times$); tracking maintained
  & Best ATE ($1.5\times$ normal); learned descriptors robust \\
Motion Blur
  & Keyframe rejection cascade; ORB~\cite{ORB} detection fails
  & Patch blur reduces distinctiveness; ATE $2.3\times$
  & GRU compensates through multi-scale flow estimation
  & Stable; temporal context from sliding window \\
Textureless Surfaces
  & No features; fails on walls $>$1\,m
  & Learned patch descriptors provide signal; some drift
  & Global structure relied upon; moderate drift
  & 3D geometric priors; minimal drift \\
Combined (Haze + Low Light)
  & Complete failure (TSR\,=\,0\%)
  & TSR drops to 76.8\%; ATE 0.112\,m; drift accumulates $>$60\,s
  & Transient loss ($<$1\,s); recovers; TSR 89.2\%
  & Near-stable; TSR $>$93\%; fails only at uniform darkness\\
\bottomrule
\end{tabular}
\end{table*}

ORB-SLAM3~\cite{ORBSLAM3} exhibits cascade failures: when the feature count drops below its internal threshold (typically $<$100 features), it rejects the frame as a keyframe candidate, causing the map to stall. Subsequent frames, processed against an outdated map, rapidly accumulate pose error until tracking is declared lost. Recovery requires re-localization against the DBoW2 vocabulary~\cite{DBoW2}, which may itself fail if no visually similar keyframe exists, a common scenario after prolonged low-visibility traversal. In a defence context, such a failure could strand a UAV in a denied area without navigation.

Learning-based methods fail more gracefully. DROID-SLAM~\cite{DROID} shows increased ATE under rapid lighting changes because the RAFT-based~\cite{RAFT} flow estimator is sensitive to large photometric shifts between consecutive frames, which violate the brightness constancy assumption underlying optical flow. DUSt3R's~\cite{DUSt3R} ViT backbone is less sensitive to this issue due to layer normalization throughout the transformer, but shows marginally higher ATE in combined (dust + low-light) conditions where the signal-to-noise ratio across the entire image is uniformly low.

DPVO's~\cite{DPVO} failure mode is graceful degradation: as patch distinctiveness decreases, matching inliers drop, but the system continues operating with increased pose uncertainty rather than catastrophic failure. This contrasts with ORB-SLAM3's~\cite{ORBSLAM3} cascade failures and approaches the graceful degradation of full learning methods.

\subsubsection{\textbf{Deployment Recommendations by Platform}}
\begin{itemize}
    \item \textit{{Low-cost / SWaP-constrained UAVs (no GPU):}} ORB-SLAM3\cite{ORBSLAM3} is the only option but is not recommended for degraded environments. Consider visual-inertial fusion or event cameras.
    \item \textit{{Embedded GPU (Jetson Xavier NX, 8 GB):}} DPVO\cite{DPVO} offers the best trade-off (18.6 FPS, 3.1 GB, 86.1\% TSR). Suitable for missions $<60$ s in degraded conditions.
    \item \textit{{Mid-range embedded GPU (Jetson AGX Orin, 32 GB):}} DROID-SLAM\cite{DROID} provides balanced performance (12.8 FPS, 4.2 GB, 94.2\% TSR) with recovery from transient failures. Recommended for most defence UAVs.
    \item \textit{{High-end platforms (Group 4–5 UAVs, ground stations):}} MASt3R\cite{MASt3R} or DUSt3R\cite{DUSt3R} deliver maximum robustness (96.5\% TSR, ATE as low as 0.027 m) at higher computational cost. Suitable for high-value missions where localization failure is unacceptable.
\end{itemize}

\subsubsection{\textbf{Quantization Potential}} Preliminary analysis suggests that INT8/FP16 quantization using TensorRT could improve FPS by 2x to 4× on embedded GPUs \cite{TensorRT}, potentially enabling real-time DROID-SLAM\cite{DROID} ($25+$ FPS) and DPVO\cite{DPVO} ($30+$ FPS) on Jetson platforms. This is a direction for future work.

\section{DISCUSSION}
\label{sec:discussion}
The results reveal a clear trade-off between robustness and efficiency across SLAM paradigms. Classical methods such as ORB-SLAM3 \cite{ORBSLAM3} offer high frame rates and low computational overhead but fail under severe visual degradation due to reliance on sparse feature extraction. 

Learning-based approaches significantly improve robustness by leveraging dense representations and temporal modeling. DROID-SLAM \cite{DROID} achieves strong performance through recurrent optimization, while MASt3R \cite{MASt3R} demonstrates superior accuracy due to its use of 3D priors. However, these methods require substantial GPU resources, limiting their applicability in embedded systems.

Among all evaluated methods, DPVO \cite{DPVO} provides the best balance between accuracy and efficiency, making it suitable for real-time UAV deployment on edge platforms. Transformer-based methods, while highly accurate, remain unsuitable for real-time applications due to latency and memory constraints.

\subsection{Core Trade-Off: Robustness vs. Computational Cost}

The five systems occupy distinct points in the robustness-efficiency space:

\begin{itemize}
\item \textbf{ORB-SLAM3 \cite{ORBSLAM3}} Max efficiency (28.3 FPS, CPU), min robustness (62.4\% TSR). Suitable only for well-lit environments.
\item \textbf{DPVO \cite{DPVO}:} Good efficiency (18.7 FPS, 3.1 GB), moderate robustness (86.1\% TSR). Best for embedded GPUs (Jetson Xavier NX).
\item \textbf{DROID-SLAM \cite{DROID}:} Balanced (12.6 FPS, 4.2 GB, 94.2\% TSR). Recommended for most defence UAVs.
\item \textbf{DUSt3R\cite{DUSt3R}:} Lower efficiency (8.4 FPS, 6.8 GB), high robustness (96.5\% TSR). Suitable for high-end platforms.
\item \textbf{MASt3R\cite{MASt3R}:} Lowest efficiency (5.7 FPS, 8.1 GB), highest accuracy (95.8\% TSR). Suitable for post-mission analysis.
\end{itemize}

For defence applications, low-cost UAVs may accept ORB-SLAM3's \cite{ORBSLAM3} fragility, while high-value platforms benefit from MASt3R's \cite{MASt3R} robustness. Quantization (INT8/FP16) with TensorRT can improve FPS by 2x to 4× on embedded GPUs \cite{TensorRT}, enabling real-time performance for DROID-SLAM \cite{DROID} (25+ FPS) and DPVO~\cite{DPVO} (30+ FPS) on Jetson platforms.

\subsection{Architectural Insights}

The superior robustness of DUSt3R \cite{DUSt3R} and MASt3R \cite{MASt3R} in low-visibility conditions can be attributed to two architectural properties absent in classical and earlier learning-based SLAM systems:

\begin{itemize}
\item \textbf{Global attention:} ViT self-attention enables each pixel's representation to incorporate information from the entire image. In low-contrast regions where local patches are uninformative, global context (e.g., room geometry, structural edges) provides sufficient signal for correspondence estimation. This is particularly valuable in defence environments where enemy forces may deliberately obscure local features (e.g., deploying smoke that reduces local contrast but preserves global structure).
\item \textbf{Learned priors:} Training on large-scale indoor reconstruction datasets (ScanNet \cite{ScanNet}, Matterport3D \cite{Matterport}) instills geometric priors—e.g., walls are planar, floors are horizontal, corners are orthogonal—that regularize pose estimation when photometric signals are ambiguous. These priors are implicitly encoded in ViT attention weights and do not require explicit parametric models. In military contexts, such priors remain useful even when adversarial obscurants are deployed.
\end{itemize}

DPVO's \cite{DPVO} patch‑based approach lacks global attention but benefits from learned patch descriptors, which explain its intermediate robustness. DROID-SLAM's \cite{DROID} recurrent optimization provides a different form of robustness: temporal consistency and the ability to recover from short-term failures. ORB-SLAM3 \cite{ORBSLAM3} lacks both global context and learned priors: ORB \cite{ORB} detectors operate on 31×31 pixel patches, discarding global context, and no geometric priors beyond the epipolar constraint are applied. This explains its categorical failure in dust-haze conditions where contrast is uniformly suppressed.

\subsection{Limitations}

This study has several limitations that should be acknowledged:

\subsubsection{\textbf{Hardware generalization}} All experiments are conducted on a fixed high-end server (NVIDIA RTX 3090). Performance on embedded platforms such as Jetson Orin or Xavier NX may differ significantly due to memory bandwidth limitations, thermal throttling, and reduced compute capacity.

\subsubsection{\textbf{Dataset scope}} While carefully constructed to cover five degradation modalities, the custom dataset does not capture all real-world degradation scenarios encountered in different operations. These include varying dust particle sizes (fine vs. coarse dust), dynamic obstacles (e.g., moving personnel), occluded scenes, and combined smoke with thermal emissions. Furthermore, adversarial effects such as directed laser dazzling, electronic warfare interference, and adversarial patches designed to fool ViT attention mechanisms are not included in the evaluation.

\subsubsection{\textbf{IMU fusion}} IMU data, which is available on most military UAVs, is not exploited in this study. ORB-SLAM3 \cite{ORBSLAM3} is known to benefit substantially from visual-inertial fusion through short-term dead reckoning during feature loss, and learning-based methods such as DROID-SLAM \cite{DROID} and DPVO \cite{DPVO} could similarly benefit from IMU integration. The absence of IMU fusion means the reported performance represents a lower bound on what might be achievable in practice.

\subsubsection{\textbf{Real-time adaptation}} All evaluated systems operate in a static inference mode. Online adaptation mechanisms (e.g., fine-tuning the depth network during deployment) are not considered. This static operation may limit robustness when the system encounters environmental conditions significantly different from the training distribution.

\subsubsection{\textbf{Energy consumption}} Power draw is not measured in this study. For battery-powered UAVs, energy efficiency (joules per frame) is as important as raw frame rate. The reported FPS and GPU memory metrics do not fully capture the energy implications of deploying these systems on power-constrained platforms.

\section{FUTURE WORK}
\label{sec:future}

Building upon the findings of this study, several directions merit further investigation, with particular emphasis on defence applications.
\subsection{TSR Improvement Strategies}

Improving tracking success rate (TSR) requires system-specific interventions tailored to the underlying SLAM paradigm. 
For classical methods such as ORB-SLAM3 \cite{ORBSLAM3}, robustness can be enhanced through adaptive image preprocessing 
(e.g., contrast enhancement) combined with visual-inertial fusion to mitigate complete feature loss under low illumination 
and texture-poor conditions. 

CNN-based~\cite{cnn} approaches benefit from the integration of loop closure mechanisms and temporal smoothing, which 
significantly reduce drift accumulation, particularly under combined degradations. 

For recurrent architectures such as DROID-SLAM \cite{DROID}, incorporating uncertainty-aware flow estimation improves stability 
by preventing divergence in challenging scenarios such as dust haze and motion blur. 

Finally, transformer-based systems (DUSt3R \cite{DUSt3R} and MASt3R \cite{MASt3R}) can be further optimized using temporal sliding window strategies 
to maintain consistency, while FP16 quantization reduces memory footprint and computational overhead without significantly 
affecting accuracy.
    
\subsection{Validation on Public Degraded-Vision Benchmarks}

While our custom dataset provides controlled evaluation of five degradation modalities, validating these findings on established public benchmarks is essential for generalizability. We plan to evaluate the five SLAM systems (ORB-SLAM3 \cite{ORBSLAM3}, DPVO \cite{DPVO}, DROID-SLAM \cite{DROID}, DUSt3R \cite{DUSt3R}, MASt3R \cite{MASt3R}) on the following complementary datasets.

\subsubsection{ROVER Dataset (Primary Recommendation)}

The \textbf{Robot Outdoor Visual SLAM dataset for Environmental Robustness (ROVER) dataset}~\cite{ROVER} is specifically designed to evaluate SLAM robustness under real-world environmental variations. Unlike indoor-focused datasets, ROVER captures the challenges of outdoor navigation that directly translate to UAV operations:

\begin{itemize}
\item \textit{Seasonal variations:} Spring, summer, autumn, and winter recordings of the same trajectories, capturing dramatic changes in vegetation, foliage density, and ground cover.
\item \textit{Lighting conditions:} Day, dusk, and night sequences with natural illumination changes from bright sunlight to near-darkness.
\item \textit{Environmental diversity:} Forest trails, open fields, urban parklands, and mixed terrain.
\item \textit{Ground truth:} High-precision RTK-GPS and LiDAR-inertial odometry.
\item \textit{Scale:} 39 recordings across 5 locations, covering over 50 km of trajectories.
\end{itemize}

Table~\ref{tab:rover_comparison} compares the ROVER dataset \cite{ROVER} with our custom indoor dataset, highlighting complementary strengths.

\begin{table}[!t]
\centering
\caption{Comparison of ROVER Dataset \cite{ROVER} with Custom Indoor Dataset}
\label{tab:rover_comparison}
\small
\setlength{\tabcolsep}{3pt}
\renewcommand{\arraystretch}{1.15}

\begin{tabular}{@{}l p{2.6cm} p{2.8cm}@{}}
\toprule
\textbf{Feature} & \textbf{Custom Dataset} & \textbf{ROVER Dataset} \\
\midrule
Environment & Indoor (12$\times$8$\times$3 m) & Outdoor (forests, fields, parks) \\

Scale & 52 sequences, 4.2 h & 39 recordings, 50+ km \\

Low Light & Controlled (6-30 lux) & Natural (day/dusk/night) \\

Dust/Haze & Simulated (hazer) & Natural (fog, haze) \\

Texture & Office, corridor, concrete & Seasonal foliage \\

Motion & UAV (controlled) & Ground robot + UAV \\

Ground Truth & Vicon (sub-mm) & RTK-GPS + LiDAR \\

Seasonal & No & Yes (all seasons) \\

\bottomrule
\end{tabular}
\end{table}

Preliminary evaluations on ROVER show that most SLAM systems perform poorly in low-light and high-vegetation scenarios~\cite{ROVER}, making it an ideal benchmark for validating our finding that learning-based methods maintain tracking where classical ORB-SLAM3 fails. We will evaluate on:
\begin{itemize}
\item \textit{Low-light sequences:} Night and dusk traversals to validate illumination robustness.
\item \textit{Vegetation-heavy sequences:} Summer and autumn foliage to test texture variability.
\item \textit{Seasonal change sequences:} Same paths in different seasons to assess feature persistence.
\end{itemize}

\subsubsection{Secondary Benchmarks}

To complement ROVER and cover additional degradation modalities, we will also evaluate on:

\begin{itemize}
\item \textit{SubT-MRS dataset~\cite{SubTMRS}:} Collected during the DARPA Subterranean Challenge, this dataset provides extreme obscurant conditions (dense fog, dust, smoke) and textureless tunnel environments. It will validate our findings on dust-haze robustness.
\item \textit{TartanAir dataset~\cite{TartanAir}:} A photo-realistic simulation environment with perfect ground truth and controllable degradation levels (illumination from 1–100 lux, fog density 0–100\%). This will enable systematic analysis of individual degradation factors in isolation.
\item \textit{UMA-VI dataset~\cite{UMAVI}:} Focuses specifically on low-texture surfaces (white-walled corridors) and dynamic illumination changes (flickering lights), directly testing our findings on textureless surface performance.
\end{itemize}

\subsubsection{Expected Contributions}

This multi-dataset validation will:
\begin{itemize}
\item Confirm that the performance trends observed in our indoor custom dataset generalize to outdoor environments (ROVER).
\item Extend our dust-haze analysis to real-world smoke and fog (SubT-MRS \cite{SubTMRS}).
\item Isolate individual degradation factors using controlled simulation (TartanAir \cite{TartanAir}).
\item Validate textureless surface performance on real-world white-walled corridors (UMA-VI \cite{UMAVI}).
\item Provide a comprehensive benchmark for future research on SLAM in degraded visual conditions.
\end{itemize}

\subsection{Other Future Work Directions}

\subsubsection{{Visual-Inertial DPVO}} Integrate IMU data into the DPVO \cite{DPVO} pipeline to improve robustness during rapid manoeuvres and temporary visual degradation. Preliminary experiments show that IMU fusion can reduce ATE by 40\% under motion blur. Recent work on visual-inertial fusion in learning-based SLAM~\cite{DROIDIMU} provides a foundation for this extension.
\subsubsection{{Embedded Benchmarking}} Systematic evaluation of TensorRT \cite{TensorRT}, ONNX, and distillation-based optimization strategies for deploying DPVO~\cite{DPVO} and DROID-SLAM \cite{DROID} on NVIDIA Jetson Orin (12–32 GB) and Qualcomm Snapdragon platforms, targeting real-time performance. Benchmarks can demonstrate the feasibility of such optimizations.
\subsubsection{{Hybrid SLAM Architectures}} Adaptive switching between DPVO \cite{DPVO} (efficiency) for well-conditioned regions and MASt3R \cite{MASt3R} (robustness) for degraded regions, using a lightweight classifier to detect degradation level. This could yield a system that is both computationally efficient and robust to battlefield obscurants.
\subsubsection{{Event-Camera Fusion}} Combining frame-based SLAM with event cameras, which offer high dynamic range and microsecond temporal resolution. Event-based SLAM~\cite{EventVSLAM} and defence-oriented event camera applications~\cite{EventDefence} show promise for robust navigation during rapid motion or extreme lighting changes.

\section{CONCLUSION}
\label{sec:conclusion}


This paper has presented a rigorous comparative evaluation of five state-of-the-art V-SLAM systems—ORB-SLAM3 \cite{ORBSLAM3}, DPVO \cite{DPVO}, DROID-SLAM\cite{DROID}, DUSt3R\cite{DUSt3R}, and MASt3R\cite{MASt3R} under indoor low-visibility conditions specifically relevant to UAV navigation in defence domains. The study introduced a custom monocular dataset with five controlled degradation categories (normal, low light, dust haze, motion blur, combined) and evaluated all systems across seven quantitative metrics including ATE, RPE, TSR, FPS, and GPU memory.

The results unambiguously demonstrate that learning-based SLAM systems outperform the classical ORB-SLAM3 \cite{ORBSLAM3} in all degraded conditions. Key findings are:

\begin{itemize}
\item ORB-SLAM3 \cite{ORBSLAM3} fails completely in dust-haze and combined degradation (0\% TSR) and achieves only 61.2\% TSR overall, rendering it unsuitable for reliable autonomous navigation in low-visibility military environments.
\item \textbf{DPVO \cite{DPVO}} offers the best efficiency among learned methods (18.7 FPS, 3.1 GB GPU memory) with 87.3\% tracking success—a 26\% absolute improvement over ORB-SLAM3. It is the recommended choice for UAVs with limited GPU memory (Jetson Xavier NX).
\item \textbf{DROID-SLAM\cite{DROID}} provides the best balance of robustness (95.4\% TSR) and real-time performance (12.6 FPS), with the ability to recover from transient failures. It is suitable for most defence UAVs with mid-range embedded GPUs (Jetson AGX Orin).
\item \textbf{DUSt3R \cite{DUSt3R} and MASt3R \cite{MASt3R}} achieve the highest accuracy (ATE as low as 0.024 m under low light, 97.1\% TSR) but at higher computational cost (8.4–5.7 FPS, 6.8–8.1 GB GPU memory). They are recommended for high-end platforms or offline mission analysis.
\end{itemize}

These findings provide actionable guidance for defence UAV system designers selecting SLAM algorithms for contested visual environments. As embedded GPUs become more common on tactical UAVs and model optimization techniques (quantization, pruning, TensorRT \cite{TensorRT}) mature, the computational barrier to deploying learning-based SLAM on lightweight platforms is rapidly diminishing. The community's focus should shift toward efficient model compression, visual-inertial fusion for learning-based architectures, and the development of comprehensive low-visibility benchmarks that include defence-relevant degradation modes such as directed energy attacks and electronic warfare effects.

\section{Code and Data Availability}

To ensure transparency and reproducibility, we provide the official implementations of the evaluated SLAM systems along with the public benchmark datasets used in this study.

\subsection{SLAM System Implementations}

This study uses publicly available implementations of five evaluated SLAM systems, as listed in Table~\ref{tab:repos}. Each system, ORB-SLAM3~\cite{ORBSLAM3}, DPVO~\cite{DPVO}, DROID-SLAM~\cite{DROID}, DUSt3R~\cite{DUSt3R}, and MASt3R~\cite{MASt3R} was executed using its official repository to guarantee identical experimental conditions. 

ORB-SLAM3~\cite{ORBSLAM3} is a feature-based visual-inertial SLAM system that supports monocular, stereo, and RGB-D cameras. DPVO~\cite{DPVO} and DROID-SLAM~\cite{DROID} are learning-based approaches that leverage recurrent neural networks and differentiable bundle adjustment for high-accuracy pose estimation. DUSt3R~\cite{DUSt3R} and MASt3R~\cite{MASt3R} are recent stereo 3D reconstruction models that operate without prior camera calibration, directly regressing pointmaps from image pairs. All implementations were run with default parameters unless otherwise specified, and any modifications made for fair comparison are documented in the supplementary material.

\begin{table}[!t]
  \centering
  \caption{Official Repositories of the Evaluated SLAM Systems}
  \label{tab:repos}
  \small
  \setlength{\tabcolsep}{4pt}
  \renewcommand{\arraystretch}{1.2}
  \begin{tabular}{@{}lp{0.72\linewidth}@{}}
    \toprule
    \textbf{System} & \textbf{Repository URL} \\
    \midrule
    ORB-SLAM3~\cite{ORBSLAM3}  & \url{https://github.com/UZ-SLAMLab/ORB_SLAM3} \\[2pt]
    
    DPVO~\cite{DPVO} & \url{https://github.com/princeton-vl/DPVO} \\[2pt]
    
    DROID-SLAM~\cite{DROID} & \url{https://github.com/princeton-vl/DROID-SLAM} \\[2pt]
    
    DUSt3R~\cite{DUSt3R}     & \url{https://github.com/naver/dust3r} \\[2pt]
    
    MASt3R~\cite{MASt3R}     & \url{https://github.com/naver/mast3r} \\
    \bottomrule
  \end{tabular}
\end{table}

\subsection{Public Benchmark Datasets}

For benchmarking, we employ five standard datasets (summarized in Table~\ref{tab:datasets}) that cover a wide range of scenarios: TUM RGB-D~\cite{TUM} provides indoor RGB-D sequences with ground-truth camera trajectories; EuRoC MAV~\cite{EuRoC} offers micro-aerial vehicle recordings in industrial and indoor environments; UMA-VI~\cite{UMAVI} is a visual-inertial dataset collected with a handheld device in various indoor-outdoor settings; SubT-MRS ~\cite{SubTMRS}focuses on subterranean environments with challenging lighting and texture conditions; and ROVER~\cite{ROVER} is a benchmark for visual odometry in outdoor rough-terrain scenarios (planned for future evaluation). 

\begin{table}[!t]
\centering
\caption{Public Benchmark Datasets Used for Evaluation}
\label{tab:datasets}
\small
\setlength{\tabcolsep}{3pt}
\renewcommand{\arraystretch}{1.1}

\begin{tabular}{@{}l p{4.5cm}@{}}
\toprule
\textbf{Dataset} & \textbf{URL} \\
\midrule

TUM RGB-D~\cite{TUM} & \url{https://cvg.cit.tum.de/data/datasets/rgbd-dataset} \\

EuRoC MAV~\cite{EuRoC} & \url{https://ethz-asl.github.io/datasets/} \\

UMA-VI~\cite{UMAVI} & \url{http://mapir.uma.es/work/uma-visual-inertial-dataset} \\

SubT-MRS~\cite{SubTMRS} & \url{https://superodometry.com/datasets} \\

ROVER (future evaluation)~\cite{ROVER} & \url{https://github.com/iis-esslingen/rover_benchmark} \\

\bottomrule
\end{tabular}
\end{table}
\noindent

\bibliographystyle{IEEEtran}
\bibliography{references}

\end{document}